\documentclass{article}

\usepackage{arxiv}

\usepackage[utf8]{inputenc} 
\usepackage[T1]{fontenc}    
\usepackage{hyperref}       
\usepackage{url}            
\usepackage{booktabs}       
\usepackage{amsfonts}       
\usepackage{nicefrac}       
\usepackage{microtype}      
\usepackage{lipsum}

\usepackage{relsize}
\usepackage{verbatim}
\usepackage{graphicx}
\usepackage{dcolumn}
\usepackage{bm}
\usepackage{float}

\usepackage{graphics}
\usepackage{color}
\usepackage{xcolor}
\usepackage{bm}

\usepackage{soul}
\usepackage{xcolor}
\usepackage{graphicx}
\usepackage{hyperref}
\usepackage{subcaption}   
\usepackage{tikz}
\usetikzlibrary{arrows.meta,positioning,fit,calc}
\usepackage{booktabs}
\usepackage{bm}
\usepackage{mathtools}

\usepackage{amsmath,amssymb,amsfonts}

\usepackage{algorithm}
\usepackage{algpseudocode}
\algnewcommand\server{\item[\textbf{Server execution:}]}%
\algnewcommand\client{\item[\textbf{ClientUpdate($k,w$):}]}%

\usepackage{enumitem}
\setlist[itemize]{leftmargin=*}
\setlist[enumerate]{leftmargin=*}
\usepackage{amsmath,amssymb,amsfonts}
\usepackage{graphics}
\usepackage{color}
\usepackage{xcolor}
\definecolor{rev}{rgb}{0,0,0}
\definecolor{rev2}{rgb}{0,0,0}

\usepackage{array}
\newcolumntype{P}[1]{>{\centering\arraybackslash}p{#1}}
\usepackage{multirow}
\usepackage{cancel}
\usepackage[capitalise]{cleveref} 
\usepackage{enumitem}

\usepackage{cite}

\usepackage{tabularx}
\usepackage{threeparttable}
\usepackage{pifont} 
\newcolumntype{Y}{>{\centering\arraybackslash}X}

\title{SIMR-NO: A Spectrally-Informed Multi-Resolution Neural Operator for Turbulent Flow Super-Resolution}

\author{
  Muhammad Abid \\
  Department of Mechanical and Aerospace Engineering,\\
  University of Tennessee,\\
  Knoxville, TN 37996, USA.\\
  \texttt{mabid@vols.utk.edu} 
  \And
  Omer San \\
  Department of Mechanical and Aerospace Engineering,\\
  University of Tennessee,\\
  Knoxville, TN 37996, USA.\\
  \texttt{osan@utk.edu}
}

\begin{document}
\maketitle

\begin{abstract}
Reconstructing high-resolution turbulent flow fields from severely under-resolved
observations is a fundamental inverse problem in computational fluid dynamics and
scientific machine learning. Classical interpolation methods fail to recover
missing fine-scale structures, while existing deep learning approaches rely on
convolutional architectures that lack the spectral and multiscale inductive biases
necessary for physically faithful reconstruction at large upscaling factors. We
introduce the \textit{Spectrally-Informed Multi-Resolution Neural Operator}
(SIMR-NO), a hierarchical operator learning framework that factorizes the
ill-posed inverse mapping across intermediate spatial resolutions, combines
deterministic interpolation priors with spectrally gated Fourier residual
corrections at each stage, and incorporates local refinement modules to recover
fine-scale spatial features beyond the truncated Fourier basis. The proposed
method is evaluated on Kolmogorov-forced two-dimensional turbulence, where
$128\times128$ vorticity fields are reconstructed from extremely coarse $8\times8$
observations representing a $16\times$ downsampling factor. Across 201
independent test realizations, SIMR-NO achieves a mean relative $\ell_2$ error
of $26.04\%$ with the lowest error variance among all methods, reducing
reconstruction error by $31.7\%$ over FNO, $26.0\%$ over EDSR, and $9.3\%$
over LapSRN. Beyond pointwise accuracy, SIMR-NO is the only method that
faithfully reproduces the ground-truth energy and enstrophy spectra across the
full resolved wavenumber range, demonstrating physically consistent
super-resolution of turbulent flow fields.
\end{abstract}


\section{Introduction}\label{sec:intro}

The process of reconstructing high-resolution turbulent flow fields from extremely limited observational data serves as an essential inverse problem that connects the fields of computational fluid dynamics (CFD) and scientific machine learning (SciML) research. The practical world only permits researchers to observe flow patterns through spatial measurements, which are too broad to be useful for their numerical simulations and laboratory experiments and field measurements because their equipment does not provide sufficient resolution \cite{tennekes1972first}. The field of fluid dynamics uses various physical quantities, which are required to study fine-scale structures that include vortical filaments and shear layers and small-scale turbulent fluctuations. These structures control all three turbulent dynamic processes, which include energy transfer and mixing and dissipation
processes that are central to understanding turbulent dynamics
\cite{pope2000turbulent}. The process of recovering missing spatial scales from restricted observations happens to be more than just a technical feature because it serves as an essential requirement that need to perform accurate analyses and predictions about turbulent systems and their corresponding models
\cite{brunton2020mlfluid,vinuesa2022enhancing,kramer2024operatorinference}.

The solution to this problem becomes challenging because downsampling leads to permanent data loss. The high-resolution flow field becomes inaccessible because only coarse-resolution data provides visibility to high-frequency spatial details that create an inverse reconstruction problem that conforms to Hadamard's definition of ill-posedness through its dual nature of non-constant solutions, which react strongly to changes in input data \cite{engl1996regularization}. Nonlinear multiscale interactions in turbulent flows create additional ill-posedness because they create new energy while spreading existing energy across different spatial dimensions through the turbulent energy cascade \cite{kolmogorov1941local,frisch1995turbulence}. Classical interpolation techniques such as bilinear or bicubic interpolation show the ability to reconstruct smooth large-scale variations, but they cannot recover the small-scale vortical structures that create turbulence and must follow physical laws. The resulting reconstructions become excessively smooth because they constantly misrepresent fine-scale energy components that control important downstream quantities.

The scientific field of machine learning has developed new techniques that use data to create effective methods for studying intricate physical systems. The method of operator learning provides a structured approach that enables researchers to create mathematical models that describe how infinite-dimensional mappings between function spaces behave under partial differential equations (PDEs). The operation of neural operators differs from traditional neural networks because they learn to create continuous mappings between function spaces, which allow them to make predictions without needing specific dimensional inputs \cite{chen1995universal}. DeepONet \cite{lu2021deeponet} and the Fourier Neural Operator (FNO) \cite{li2020fno} have established themselves as essential frameworks that enable the development of mathematical models that simulate various physical phenomena, including fluid dynamics, climate change, and elasticity \cite{kovachki2023neuraloperator}. The main benefit of these models enables the creation of operators that function across different resolutions while maintaining the ability to learn nonlocal features that exist outside of the training set. Researchers have developed neural operator systems that now support physical constraints and irregular geometries while using advanced transformer and multiscale systems \cite{li2021pino,li2023geofno,rahman2022uno,hao2023gnot,wang2022improved} which now enable the study of complicated scientific problems.

Simultaneously, machine learning methods have increasingly been applied to turbulence modeling
and flow reconstruction. Data-driven approaches have been developed for turbulence closure
modeling \cite{ling2016invariance}, subgrid-scale modeling \cite{maulik2019subgrid},
reduced-order modeling \cite{loiseau2018constrained}, and machine-learning-accelerated CFD
\cite{kochkov2021accelerated,brunton2020mlfluid,vinuesa2022enhancing,kramer2024operatorinference},
collectively demonstrating the transformative potential of learning-based methods in
computational fluid mechanics. More specifically, a growing body of work has demonstrated that
deep neural networks can reconstruct high-resolution turbulent flow fields from coarse
observations with substantially improved accuracy relative to classical interpolation baselines
\cite{fukami2019sr,fukami2021spatiotemporal,liu2020turbsr,fukami2023review,fukami2023super,taira2025machine}. The methods depend on convolutional neural network (CNN) architectures, which were developed to solve image super-resolution problems while using residual learning and dense connectivity and adversarial training to enhance visual quality. The models successfully capture local spatial relationships; however, they fail to recognize the fundamental spectral and multiscale patterns that define turbulent flows. The limitation becomes more significant when using higher upscaling factors because they need to restore accurate fine-scale details, which presents the most difficult task.

The existing super-resolution methods face their first obstacle because they use a single-stage design, which requires the model to create complete high-resolution results from coarse input data through one single training session. The model needs to learn two distinct tasks when upscaling with an $8\times$ to $128\times$ factor because it should create both the large flow patterns and the missing small flow patterns through one single mapping process. The process creates an optimization path that contains extreme nonlinear characteristics and results in two outcomes: it either removes high-frequency details or it fails to control energy distributions according to physical laws \cite{dong2016srcnn,lim2017edsr}. Flow behavior in turbulence encounters major problems because pointwise field values do not contain all the essential physical information, which also requires energy spectra and enstrophy distributions and modal energy content and coherent vortical organization to be included
\cite{kraichnan1967inertial,boffetta2012twodim,taira2020modal}. The super-resolution method, which produces low mean squared error results but fails to accurately show energy spectra and vortical topology, cannot provide valid results for future analysis and modeling work \cite{bao2023deeplearning}.

The current study develops a new architectural design, which it names 
\textit{Spectrally-Informed Multi-Resolution Neural Operator} (SIMR-NO), to solve 
existing challenges. The core idea is to reformulate turbulence super-resolution as a 
hierarchical inverse operator learning problem, decomposing the reconstruction operator 
into a sequence of progressively refined stages rather than recovering all missing spatial 
scales through a single monolithic mapping. At each stage, a deterministic bicubic 
interpolation prior provides a smooth structural estimate, upon which the learned neural 
operator applies a spectral residual correction to restore lost high-frequency content. 
Each stage produces an output whose resolution matches its input query, enabling strictly 
incremental refinement that mirrors the hierarchical energy cascade of turbulent flows 
\cite{lai2017lapsrn,he2016deep}. This stands in direct contrast to single-stage 
architectures, which must simultaneously recover all spatial scales through one highly 
nonlinear mapping, and to U-Net-type architectures, which couple encoder-decoder 
representations at fixed resolution levels and lack flexibility to generalize across 
arbitrary scales. The hierarchical formulation instead yields fewer learnable parameters, 
reduced FLOPs per query, and natural generalization to unseen resolutions, connecting to 
the well-established advantages of classical multigrid solvers 
\cite{trottenberg2000multigrid} and progressive training strategies 
\cite{karras2018progressive,bengio2009curriculum}, where coarse-to-fine decomposition 
accelerates convergence relative to monolithic approaches.

The primary architectural advancement of SIMR-NO structural design emerges through its implementation of spectrally gated convolution layers, which manage learned Fourier multipliers according to their radial wavenumber. The gated mechanism allows the model to use learned weights for all spectral modes during its operation since standard FNO layers require uniform weight application. The system detects spatial patterns that have different size scales in turbulent flows because it uses an inductive bias that the system implements. The energy distribution of turbulent flows follows low-frequency patterns while their enstrophy moves to higher-frequency patterns according to \cite{kraichnan1967inertial,boffetta2012twodim,leith1968diffusion}. The combination of hierarchical reconstruction with spectrally informed operator layers establishes a unified framework in SIMR-NO, which connects operator learning, multiscale decomposition, and spectral modeling methods to achieve accurate turbulence super-resolution results.

The proposed architecture is evaluated on a challenging benchmark that uses Kolmogorov-forced two-dimensional turbulence as its testing standard because this method enables super-resolution evaluation through its extensive multiscale attributes combined with precise spectral statistical measurements
\cite{boffetta2012twodim,san2015stabilized,taira2025machine}. The vorticity fields that study exist on a spatial grid that uses $128 \times 128$ dimensions, and these fields get reconstructed from extremely coarse $8 \times 8$ observations, which represent a $16\times$ upscaling factor. The problem becomes more difficult because the downsampling ratio exceeds standard limitations found in image super-resolution research, and it needs a model that can create realistic small-scale details instead of providing smooth detail interpolation.
The main contributions of this work are summarized as follows:

\begin{itemize}
    \item \textbf{Hierarchical inverse operator learning formulation.} We reformulate
    turbulence super-resolution as a multi-stage inverse problem, decomposing the
    reconstruction operator across intermediate spatial resolutions to progressively
    recover multiscale turbulent structures.

    \item \textbf{SIMR-NO architecture.} We introduce a novel multi-resolution neural
    operator that combines deterministic interpolation priors with spectrally informed
    residual correction operators at each stage of the reconstruction hierarchy.

    \item \textbf{Spectrally gated convolution.} We propose a radial wavenumber-based
    gating mechanism for Fourier convolution layers that endows the model with scale-aware
    inductive biases aligned with turbulent energy spectra.

    \item \textbf{Improved reconstruction accuracy.} Our results show that we achieve better quantitative results than the strong baselines FNO, EDSR, and LapSRN across all three evaluation metrics, which include MSE, SSIM, PSNR, and relative $\ell_2$ error metrics that we tested on 201 independent test cases.

    \item \textbf{Physical fidelity.} The study demonstrates that SIMR-NO research exceeds pointwise accuracy by delivering better reconstruction results that maintain energy spectra and enstrophy distributions and modal energy content better than all other methods tested. The study demonstrates that the method is appropriate for analysis of turbulence which has physical significance.
\end{itemize}

The remainder of the paper is organized as follows.
Section~\ref{sec:related_work} reviews related work on neural operators and turbulence
super-resolution. Section~\ref{sec:dataset} describes the physical dataset and the
mathematical formulation of the reconstruction problem. Section~\ref{sec:method} presents
the SIMR-NO architecture in detail. Section~\ref{sec:experiments} describes the
experimental setup and evaluation protocol. Section~\ref{sec:results} reports quantitative
and qualitative reconstruction results. Finally,
Section~\ref{sec:conclusion} concludes the paper and outlines directions for future work.

\section{Related Work}
\label{sec:related_work}

\subsection{Neural Operators and Operator Learning}

Operator learning serves as an effective method for creating mathematical models that describe infinite-dimensional mappings between function spaces. The primary goal of this research is to develop a unified operator that can handle all different variations of a partial differential equation. The approach used here differs from standard neural networks, which require separate training for every new input because they process fixed-dimensional vector data. By learning mappings between function spaces
directly, neural operators enable discretization-agnostic inference and
resolution-independent generalization \cite{chen1995universal},
properties that are especially valuable in scientific computing where simulation data may
be available at multiple resolutions and where generalization to unseen discretizations is
essential for practical deployment \cite{kovachki2023neuraloperator}.

The first functional method that developed this field of research was DeepONet
\cite{lu2021deeponet}, which created the necessary theoretical basis to use its branch-trunk network structure for neural operator approximation. The method showed that operators can be accurately approximated by using computational methods that follow the universal approximation theorem, thus enabling operators to be learned from data in various types of PDE systems. The Fourier Neural Operator (FNO) developed by \cite{li2020fno} establishes a scalable system that uses Fourier domain parameters to create global convolution kernel functions. FNO models use Fourier multipliers to train their systems, which enables them to model long-range spatial dependencies and nonlocal interactions through truncated spectral modes. The model shows its highest performance level when it handles physical systems that use elliptic or parabolic PDEs, which produce smooth solutions. The FNO model, together with its different variations, has demonstrated successful performance across several forward modeling tasks, which include simulating Navier-Stokes flows and estimating Darcy permeability and predicting global weather patterns \cite{pathak2022fourcastnet}.

Researchers established a deeper understanding of neural operators through their studies, which used existing theoretical foundations as their basis for development. Kovachki et al.\ \cite{kovachki2023neuraloperator} developed a mathematical framework that integrates all operator learning approximation techniques used in Banach spaces, while the framework specifies the precise conditions that allow neural operators to model all continuous nonlinear operators. Theoretical progress has established how the adjoint function operates within operator learning \cite{boulle2024adjoint}. The research shows precise algorithmic performance metrics, which include both convergence rates and generalization limits \cite{kovachki2024operatorlearning}. The recent advancements in neural operators have established their mathematical foundation, which matches the standards of traditional numerical techniques used in finite element analysis and spectral methods. The established foundation of this system identifies its correct application for essential scientific research.

The developed architectural extensions for neural operators, which now provide enhanced flexibility together with improved capabilities to manage complex physical systems. The deformation-based FNO \cite{li2023geofno} operator learning system performs irregular domain learning through its ability to learn deformation patterns that transform all geometric shapes into standard grid formats. The study of physics-informed neural operators \cite{li2021pino} uses partial differential equation residuals as soft training constraints, which help them achieve better physical accuracy while learning from small datasets. U-shaped neural operator architectures
\cite{rahman2022uno} use encoder-decoder structures with skip connections to create a system that automatically extracts multiscale features from different spatial levels.
The GNOT operator architecture from \cite{hao2023gnot} uses attention mechanisms to model nonlocal dependencies that exist between different physical domains and unstructured query points. Improved FNO variants \cite{wang2022improved}
have addressed spectral aliasing and expressiveness limitations of the original
architecture through modified normalization and operator parameterizations. The use of neural operators for inverse reconstruction tasks, which require recovering hidden detailed fields from incomplete coarse observations, remains less developed than its application in forward modeling. The current research specifically aims to fill this research gap through its examination of turbulence super-resolution, which serves as a fundamental and essential practical example of inverse problems.

\subsection{Machine Learning for Turbulence Reconstruction}

Researchers have started using machine learning techniques for turbulence research, which includes turbulence closure models and reduced-order models and flow state estimation and data-driven reconstruction methods. The work of Ling et al. \cite{ling2016invariance} showed that neural networks that contain Galilean invariant features can predict Reynolds stress anisotropy tensors with greater accuracy than standard linear eddy-viscosity closure models. The study by Maulik and San \cite{maulik2019subgrid} demonstrated that
two-dimensional turbulence modeling could be improved through the use of convolutional networks, which developed subgrid-scale models that surpassed traditional Smagorinsky-type eddy-viscosity methods. The research work after this conducted an in-depth assessment of architectural design alternatives, which studied their impact on physical accuracy preservation during different turbulent flow situations \cite{maulik2020subgrid}. Beck et al.\
\cite{beck2019deeples} created a three-dimensional large-eddy simulation
through data-driven closure methods, which they used with deep residual networks to demonstrate that learned subgrid models could function as substitutes for conventional models. The training process of learned turbulence models now achieves better physical consistency through physics-informed methods, which use governing equation residuals as training objectives \cite{wu2018physics}. The researchers demonstrated that data-driven methods could accurately model turbulence through their collaborative work, which resulted in the advancement of machine learning-based CFD acceleration research \cite{kochkov2021accelerated, brunton2020mlfluid, vinuesa2022enhancing, kramer2024operatorinference}.

The research field has grown increasingly active because researchers use deep learning techniques to perform turbulence super-resolution, which reconstructs high-resolution flow fields from coarse observations. The research by Fukami et al.\ \cite{fukami2019sr} proved that convolutional neural networks trained on paired coarse-to-fine flow data can achieve better results than bicubic interpolation when recovering high-resolution turbulent flow structures from low-resolution inputs. Their work established a benchmark pipeline and evaluation methodology that has since been widely adopted and extended by the community. The subsequent research demonstrated that spatial-temporal super-resolution evaluation required deep learning models to retrieve complete turbulent dynamic information, which was restricted to spatial and temporal measurements \cite{fukami2021spatiotemporal}. The single-snapshot setting \cite{fukami2024singlesnapshot,fukami2024single} addresses the practically important scenario where time-series data are unavailable, which requires the model to create detailed structural reconstructions from only one basic observation that lacks time information. The comprehensive review presents contemporary research developments in machine learning-based flow reconstruction and discovers existing research gaps in the field\cite{fukami2023review,fukami2023super,fukami2025observable}.

The advancement of turbulence super-resolution methods has advanced through two main approaches which include physics-based methods and generative techniques. Gao et al.\ \cite{gao2021picnnsr} developed physics-informed CNNs which employ Navier-Stokes residuals to establish training loss functions that enable super-resolution without requiring matched high-resolution training data. The study from \cite{kim2021unsupervised} shows that unsupervised methods can learn super-resolution mappings by using statistical characteristics of turbulent flow patterns without needing high-resolution paired data for training. Researchers have developed graph neural network architectures that \cite{sanchezgonzalez2020meshgraphnets,ozawa2024spatial,stahl2025sparse} enable flow reconstruction through irregular and unstructured mesh systems, which provide more geometric flexibility than structured convolutional methods. The research area now investigates three-dimensional turbulence reconstruction, which uses two-dimensional observation data
\cite{yousif2023reconstruct3d}while studying subgrid-scale super-resolution for large-eddy simulation pipelines
\cite{pang2024sgssr,yu2025three,wu2026super,wu2025generalizable,ye2025enhancing} and implementing loss functions that address spectral and dynamical discrepancies between reconstructed and target fields
\cite{page2025dynamicsloss,page2025super}. The probabilistic and diffusion-based methods work for uncertainty estimation in super-resolved fields by delivering confidence  assessments that accompany their point reconstruction outputs \cite{price2023probabilistic}. This feature serves important benefits that assist in analyzing turbulent flow applications.

Most current turbulence super-resolution solutions exhibit a fundamental design constraint, which requires them to use convolutional neural networks that were developed for image restoration work. These networks use unchanged spatial filtering windows to detect nearby spatial patterns in data, but they fail to represent the complete atmospheric disturbance behavior that exists in turbulent flow. Standard CNNs have no built-in mechanism to distinguish energy-containing large-scale modes from enstrophy-cascading small-scale modes \cite{kraichnan1967inertial,boffetta2012twodim}, nor to adaptively emphasize spectral corrections at physically relevant wavenumber ranges. The limitation becomes more important when the system applies large upscaling factors because the system must reconstruct accurate fine-scale structures instead of creating smooth large-scale variations. The standard mean squared error training objectives fail to penalize spectral mismatches, which enables a model to achieve low pixel-level error while producing physically incorrect energy distributions \cite{bao2023deeplearning,taira2025machine}. The present work directly addresses both of these limitations by embedding spectral awareness into the reconstruction operator through Fourier-domain gating mechanisms and by adopting a hierarchical reconstruction strategy that progressively recovers fine-scale content across spatial scales.

\subsection{Deep Learning for Image Super-Resolution}

The research field of image super-resolution functions as the fundamental basis that develops scientific super-resolution methods through its provision of architectural designs and training methods and evaluation standards that are used in the field of fluid mechanics. The SRCNN \cite{dong2016srcnn} system proved that end-to-end learned super-resolution systems based their operation on shallow convolutional networks. The system achieved significant performance gains because its three-layer design processed paired low- and high-resolution image patches during training. The researchers conducted their studies to build deeper networks, which achieved better results through the application of residual learning methods from \cite{kim2016vdsr} and the combination of increased network depth with skip connections from \cite{he2016deep}. The researchers used recursive unfolding techniques from \cite{kim2016drcn,tai2017drrn} to enhance parameter efficiency because the method allowed weight sharing across different reconstruction processes. The research used two different methods of sub-pixel upsampling, which improved its inference performance through efficient feature extraction in low-resolution space followed by one learned upsampling process \cite{dong2016fsrcnn,
shi2016espcn}.

The present research work utilizes multiple architectural designs that were developed during this historical time frame because these designs use multiscale and progressive design methods. The system created by LapSRN \cite{lai2017lapsrn} uses a Laplacian pyramid for progressive upsampling, which allows the system to reconstruct an image through multiple resolution stages instead of processing it at once. The pyramid system contains multiple levels, each of which uses a small convolutional branch to calculate the difference between the current upscaled prediction and the actual result that exists at that particular level. The organization of this system into hierarchical structures enables users to learn large-magnification super-resolution because the system breaks down complex tasks into easier components that directly support the multiple-resolution decomposition method used in SIMR-NO. EDSR \cite{lim2017edsr} proved that deep residual networks achieve better performance without batch normalization because the normalization layers remove essential scale information needed for accurate reconstruction. EDSR serves as a strong convolutional baseline in our experiments. RDN \cite{zhang2018rdn} and DBPN \cite{haris2018dbpn} used dense connectivity and iterative back-projection mechanisms to enhance feature reuse and multi-scale representation learning, which resulted in better recovery of high-magnification fine-grained structures.

The development of image super-resolution has progressed through the recent introduction of attention-based and generative approaches. RCAN \cite{zhang2018rcan} channel attention networks, proved that using feature channel weights according to their information value enables the creation of high-quality reconstruction results that efficiently recover detailed textures. The ESRGAN \cite{wang2018esrgan} adversarial approach changed its optimization process to focus on training a discriminator that distinguishes between super-resolved and actual high-resolution images, which resulted in sharper visual output but decreased pixel accuracy. The SwinIR \cite{liang2021swinir} transformer restoration model used self-attention mechanisms from \cite{vaswani2017attention} to develop advanced reconstruction capabilities that use long-distance spatial information that convolutional systems with restricted field of view limitations cannot track. The diffusion model-based super-resolution method in \cite{saharia2022image} introduces stochastic iterative refinement as a systematic approach that produces multiple accurate high-quality reconstructions that show the uncertainty of high-frequency details, which is essential for turbulence studies that deal with unpredictable fine-scale patterns that can only be observed through basic measurements. The efficient cascaded architectures from \cite{ahn2018carn} together with the task-adaptive frameworks from \cite{kong2021classsr}, enable achieving better super-resolution results under their current system resource constraints while solving actual implementation challenges.

Architectural advancements introduce essential design concepts, which include hierarchical reconstruction and residual correction and attention-based feature weighting and long-range dependency modeling. The process of turbulence super-resolution operates through a fundamental distinction from image restoration because it requires physical correctness to achieve proper results. The reconstructed vorticity field displays a clear appearance yet fails to meet perceptual accuracy because it fails to generate the correct energy spectrum and enstrophy distribution together with the specific vortical structure \cite{bao2023deeplearning}. Standard image quality metrics such as SSIM and PSNR measure structural and luminance similarity, but they do not assess spectral inconsistencies or modal energy errors or violations of the governing conservation principles. The scientific super-resolution community has started to acknowledge the fundamental misalignment between image restoration objectives and physical fidelity requirements because this issue directly affects their research work \cite{page2025dynamicsloss,page2025super,stengel2020adversarial}. The present study develops architectural designs that use spectral inductive biases to model turbulent flow physics according to its natural behavior.
\subsection{Positioning of the Present Work}

The SIMR-NO framework combines three different research areas, which include neural operator learning and turbulence super-resolution and hierarchical image reconstruction. The system uses common research methods to build a system that unifies their different strengths while solving their separate research problems. The SIMR-NO system uses a hierarchical multi-resolution framework that enables the system to resolve the inverse mapping problem through multiple spatial resolution levels. FNO systems use standard architecture designs to handle forward modeling tasks that require matching input data with output data that uses similar resolution grids. The architecture of FNO systems uses one operational design to handle tasks that require gradual recovery of detailed elements that appear in high-definition super-resolution. The hierarchical decomposition in SIMR-NO reduces the effective learning complexity at each stage, which results in better optimization conditions because the system divides difficult global mapping into simpler residual corrections. The system enables gradual development of multiscale corrections that match the turbulent energy cascade \cite{kolmogorov1941local,frisch1995turbulence}.

The architectural design of SIMR-NO, which uses neural operator systems and their Fourier-domain gating system for spectral data processing, exists as a distinct approach to super-resolution modeling when compared with EDSR and LapSRN. The LapSRN method follows a method of progressive reconstruction, yet its entire system depends on local convolutional processes that use unchanging receptive field dimensions, which fail to capture the complete spectral characteristics of turbulent flow patterns. SIMR-NO replacement of convolutional residual blocks at all pyramid levels uses spectrally gated Fourier operator layers that provide nonlocal scale-aware corrections that match the physical wavenumber-dependent statistical behavior of two-dimensional turbulence \cite{san2015stabilized,boffetta2012twodim}. SIMR-NO uses built-in physical architectural features to create its system design, but physics-informed methods use soft penalty terms to implement their PDE constraints in training loss functions according to \cite{li2021pino,gao2021picnnsr,wu2018physics}. The spectrally gated convolution system
provides precise frequency-based correction capabilities that match the
two-dimensional turbulence spectral characteristics of its dual cascade process that
separates energy at large scales and enstrophy at small scales \cite{kraichnan1967inertial,leith1968diffusion} because it does not need access to
the governing Navier-Stokes equations during training. The SIMR-NO system functions
in situations where practitioners lack knowledge of the base PDE and face challenges to solve it and when experimental data replaces numeric simulation data.

SIMR-NO functions as a deterministic operator learning system that operates according to established scientific principles while maintaining accurate spectral measurements, and it does not emulate random behavior  \cite{saharia2022image,price2023probabilistic}. While generative models can create multiple possible outcomes that match their basic data input, they need higher processing power during their testing phase because they do not maintain spectral consistency with actual turbulence patterns. SIMR-NO provides a fast single-pass reconstruction process that effectively maintains both energy spectra and enstrophy distribution, thus making it suitable for use in large data assimilation and simulation speed enhancement tasks. The design choices of SIMR-NO create a system that combines physical principles with architectural design to achieve better turbulence super-resolution results through its improved reconstruction accuracy and better physical performance than current methods across all tested conditions. The research results in Sections~\ref{sec:method} and ~\ref{sec:results} demonstrate that this location between two positions results in ongoing improvements which affect standard reconstruction metrics and spectral and modal diagnostic methods used to evaluate physical performance.


\section{Dataset and Problem Formulation}
\label{sec:dataset}

The section provides an explanation of the physical system and dataset and mathematical formulation, which serves as the foundation for the turbulence super-resolution research described in this study. The researchers used inverse operator learning to develop a solution that enables them to reconstruct high-resolution vorticity fields from their extremely limited coarse dataset. The physical system is two-dimensional incompressible turbulence
that uses Kolmogorov forcing as a standard benchmark to test their data-driven flow reconstruction methods because it produces multiple spectral patterns and has established statistical characteristics \cite{boffetta2012twodim,kraichnan1967inertial,san2015stabilized}.

\subsection{Governing Equations and Physical Setup}

The underlying physical system is the two-dimensional incompressible Navier-Stokes equation
written in vorticity-streamfunction form. Let $\Omega = (0, 2\pi)^2$ denote a periodic
spatial domain, and let $\omega : \Omega \times [0,T] \rightarrow \mathbb{R}$ denote the
scalar vorticity field. The evolution of $\omega$ is governed by

\begin{equation}
    \frac{\partial \omega}{\partial t}
    + \mathbf{u} \cdot \nabla \omega
    = \nu \Delta \omega + f(x, y),
    \qquad (x,y) \in \Omega,\quad t \in [0,T],
    \label{eq:vorticity}
\end{equation}

where $\nu > 0$ denotes the kinematic viscosity and $f : \Omega \rightarrow \mathbb{R}$
is an external body forcing. The incompressible velocity field
$\mathbf{u} = (u, v) : \Omega \times [0,T] \rightarrow \mathbb{R}^2$ satisfies the
divergence-free constraint $\nabla \cdot \mathbf{u} = 0$ and is recovered from the
vorticity through the streamfunction $\psi : \Omega \times [0,T] \rightarrow \mathbb{R}$
via the kinematic relations

\begin{equation}
    \mathbf{u} = \nabla^{\perp} \psi
    \coloneqq (-\partial_y \psi,\; \partial_x \psi),
    \qquad
    \omega = \Delta \psi,
    \label{eq:streamfunction}
\end{equation}

where the second relation is the vorticity-streamfunction Poisson equation. Given
$\omega$, the stream-function is recovered by solving $\Delta \psi = \omega$ subject to
periodic boundary conditions, which is accomplished efficiently in Fourier space via

\begin{equation}
    \widehat{\psi}_{k_x, k_y}
    = -\frac{\widehat{\omega}_{k_x, k_y}}{k_x^2 + k_y^2},
    \qquad (k_x, k_y) \neq (0, 0),
    \label{eq:poisson_fourier}
\end{equation}

and $\widehat{\psi}_{0,0} = 0$ by convention. The velocity field is then obtained by
applying $\nabla^{\perp}$ in Fourier space, ensuring exact satisfaction of the
incompressibility constraint at the discrete level. To sustain a statistically stationary turbulent state, the system is driven by
Kolmogorov forcing \cite{boffetta2012twodim}, defined as

\begin{equation}
    f(x, y) = A \cos(k_f y),
    \qquad k_f = 4,
    \label{eq:forcing}
\end{equation}

where $A > 0$ controls the forcing amplitude and $k_f$ denotes the injection wavenumber.
This forcing injects kinetic energy into the system at a single spatial scale corresponding
to wavenumber $k_f = 4$, producing the characteristic large-scale banded structures in the
vorticity field that are widely used as a benchmark for turbulence modeling and
reconstruction studies. The deterministic and spatially simple nature of Kolmogorov forcing
allows the system to reach a statistically stationary state in which energy input from the
forcing is balanced by viscous dissipation, enabling the generation of long, stationary
time series of turbulent snapshots suitable for supervised learning.

\subsection{Spectral Representation and Energy Statistics}

Because the computational domain $\Omega = (0, 2\pi)^2$ is doubly periodic, the vorticity
field admits a Fourier series representation

\begin{equation}
    \omega(x, y, t)
    = \sum_{(k_x, k_y) \in \mathbb{Z}^2}
    \widehat{\omega}_{k_x, k_y}(t)\,
    e^{i(k_x x + k_y y)},
    \label{eq:fourier_expansion}
\end{equation}

where the Fourier coefficients are given by

\begin{equation}
    \widehat{\omega}_{k_x, k_y}(t)
    = \frac{1}{(2\pi)^2}
    \int_{\Omega}
    \omega(x, y, t)\,
    e^{-i(k_x x + k_y y)}\, dx\, dy.
    \label{eq:fourier_coefficients}
\end{equation}

The isotropic energy spectrum $E(k)$ is defined by averaging the modal kinetic energy over
shells of constant radial wavenumber $k = \|\mathbf{k}\|_2 = \sqrt{k_x^2 + k_y^2}$,

\begin{equation}
    E(k)
    = \frac{1}{2} \sum_{\substack{(k_x, k_y) \in \mathbb{Z}^2 \\ \|\mathbf{k}\|_2 \approx k}}
    \left|\widehat{\mathbf{u}}_{\mathbf{k}}\right|^2,
    \label{eq:energy_spectrum}
\end{equation}

where $\widehat{\mathbf{u}}_{\mathbf{k}}$ denotes the Fourier coefficient of the velocity
field at wavenumber $\mathbf{k} = (k_x, k_y)$. In two-dimensional turbulence, the dual
cascade mechanism \cite{kraichnan1967inertial,leith1968diffusion} predicts that energy
undergoes an inverse cascade toward large scales (small $k$) while enstrophy
$\mathcal{Z} = \frac{1}{2}\|\omega\|_{L^2(\Omega)}^2$ cascades forward toward small scales
(large $k$). The enstrophy spectrum is correspondingly defined as

\begin{equation}
    \mathcal{E}(k)
    = \frac{1}{2} \sum_{\substack{(k_x, k_y) \in \mathbb{Z}^2 \\ \|\mathbf{k}\|_2 \approx k}}
    \left|\widehat{\omega}_{\mathbf{k}}\right|^2
    = k^2 E(k),
    \label{eq:enstrophy_spectrum}
\end{equation}

using the relation $|\widehat{\omega}_{\mathbf{k}}|^2 = k^2 |\widehat{\mathbf{u}}_{\mathbf{k}}|^2$.
These spectral quantities serve as the primary diagnostics for evaluating the physical
fidelity of super-resolved flow fields, since they quantify the distribution of kinetic
energy and enstrophy across spatial scales in a way that pointwise error metrics such as
MSE cannot capture.

\subsection{Dataset and Discretization}

The dataset consists of $N = 1001$ temporally separated snapshots of the statistically
stationary vorticity field, generated by direct numerical simulation of the forced
Navier-Stokes system \eqref{eq:vorticity}--\eqref{eq:forcing} on the periodic domain
$\Omega$. Each snapshot is discretized on a uniform Cartesian grid with $n_h = 128$ points
per spatial dimension, yielding a high-resolution vorticity sample

\begin{equation}
    \omega^{128} \in \mathbb{R}^{128 \times 128}.
    \label{eq:hr_sample}
\end{equation}

The simulation continues until statistical stationarity is established through the convergence of both time-averaged energy spectrum and enstrophy measurements. After the statistical stationarity point, the team will take time-separated snapshots to achieve approximate statistical independence. The
resulting dataset contains a rich mixture of large-scale coherent vortical structures driven
by the Kolmogorov forcing and fine-scale turbulent fluctuations generated by the nonlinear
advection term, spanning a broad range of spatial scales across the resolved wavenumber range
$1 \leq k \leq k_{\max} = 64$. 

The dataset is partitioned into a training set of $N_{\mathrm{train}} = 800$ snapshots and
a test set of $N_{\mathrm{test}} = 201$ snapshots. All models are trained exclusively on
the training set and evaluated on the held-out test set, with no overlap between the two
partitions. This split ensures that reported metrics reflect generalization to unseen
turbulent realizations rather than in-sample reconstruction performance.

\subsection{Measurement Model and Inverse Problem Formulation}

In the setting considered here, the high-resolution field $\omega^{128}$ is not directly
observed. Instead, only a severely under-resolved coarse measurement is available, obtained
by applying a spatial restriction operator

\begin{equation}
    a^{r_c} = \mathcal{M}_{128 \rightarrow r_c}(\omega^{128})
    \in \mathbb{R}^{r_c \times r_c},
    \label{eq:downsampling}
\end{equation}

where $\mathcal{M}_{128 \rightarrow r_c}$ denotes uniform subsampling from the $128 \times
128$ grid to a coarser $r_c \times r_c$ grid. In this work, we fix $r_c = 8$, corresponding
to a $16\times$ linear downsampling factor, so that

\begin{equation}
    a^{8} \in \mathbb{R}^{8 \times 8}.
    \label{eq:coarse_obs}
\end{equation}

The downsampling operator $\mathcal{M}_{128 \rightarrow 8}$ retains only the low-frequency
content of the vorticity field, effectively discarding all Fourier modes with wavenumber
$k > k_c = r_c / 2 = 4$. Since the physically meaningful fine-scale structures of turbulence
reside precisely in the high-wavenumber modes $k_c < k \leq k_{\max}$, the measurement
process is highly destructive and the inverse reconstruction problem

\begin{equation}
    \text{find } \omega^{128} \text{ such that }
    \mathcal{M}_{128 \rightarrow 8}(\omega^{128}) = a^{8}
    \label{eq:inverse_problem}
\end{equation}

is severely ill-posed in the sense of Hadamard \cite{engl1996regularization}: infinitely
many high-resolution fields are consistent with the coarse observation $a^8$, and the
solution is highly sensitive to perturbations in the measurements. To provide a physically informed initialization for the learned reconstruction, the coarse
observation is first upsampled to the target resolution using bicubic interpolation,

\begin{equation}
    \widetilde{\omega}^{128}
    = \mathcal{U}_{8 \rightarrow 128}(a^{8})
    \in \mathbb{R}^{128 \times 128},
    \label{eq:bicubic}
\end{equation}

where $\mathcal{U}_{8 \rightarrow 128}$ denotes the bicubic interpolation operator. The
pseudo high-resolution field $\widetilde{\omega}^{128}$ provides a smooth, consistent
estimate of the large-scale flow organization, correctly capturing the low-frequency
Fourier modes $k \leq k_c$, but introduces no new spectral content and therefore fails
to recover the fine-scale vortical structures at $k > k_c$. Quantitatively, the bicubic
interpolation error is dominated by the missing high-frequency components, which is precisely the quantity that the learned neural operator must recover.

\begin{equation}
    \omega^{128} - \widetilde{\omega}^{128}
    = \sum_{\substack{(k_x,k_y) \in \mathbb{Z}^2 \\ \|\mathbf{k}\|_2 > k_c}}
    \widehat{\omega}_{\mathbf{k}}\,
    e^{i(k_x x + k_y y)},
    \label{eq:interpolation_error}
\end{equation}

\subsection{Super-Resolution as Operator Learning}

The turbulence super-resolution problem is formalized as the task of learning an operator
$\mathcal{G}_\theta$, parameterized by neural network weights $\theta \in \Theta$, that
approximately inverts the downsampling process by mapping the pseudo high-resolution
bicubic estimate to the true high-resolution vorticity field.

\begin{equation}
    \mathcal{G}_\theta
    :\; \widetilde{\omega}^{128}
    \;\longmapsto\;
    \widehat{\omega}^{128}
    \;\approx\;
    \omega^{128},
    \label{eq:operator_learning}
\end{equation}

where $\widehat{\omega}^{128} \coloneqq \mathcal{G}_\theta(\widetilde{\omega}^{128})$
denotes the reconstructed high-resolution field. The operator $\mathcal{G}_\theta$ is
trained by minimizing the empirical mean squared error over the training set.

\begin{equation}
    \mathcal{L}(\theta)
    = \frac{1}{N_{\mathrm{train}}}
    \sum_{i=1}^{N_{\mathrm{train}}}
    \left\|
        \mathcal{G}_\theta\!\left(\widetilde{\omega}^{128}_i\right)
        - \omega^{128}_i
    \right\|^2_{L^2(\Omega)},
    \label{eq:training_loss}
\end{equation}

where $\|\cdot\|_{L^2(\Omega)}$ denotes the standard $L^2$ norm on $\Omega$, approximated
in practice by the Frobenius norm of the discretized field. The learning problem
\eqref{eq:operator_learning}--\eqref{eq:training_loss} differs from standard image
super-resolution in a physically important respect: the operator $\mathcal{G}_\theta$ must
not only achieve low pointwise error but must also reproduce the correct spectral content
of $\omega^{128}$, including the energy distribution $E(k)$ and enstrophy distribution
$\mathcal{E}(k)$ across all resolved wavenumbers. The design of SIMR-NO from Section~\ref{sec:method} uses spectral and multiscale architectural elements because it needs to achieve two goals, which are pointwise accuracy and physical fidelity. The assessment of reconstruction quality uses four different metrics which provide complementary evaluations of the $N_{\mathrm{test}} = 201$ test samples that were reserved for testing purposes. The mean squared error (MSE) and relative $L^2$ error serve as tools to assess the accuracy of pointwise reconstruction: 

\begin{equation}
    \mathrm{MSE}
    = \frac{1}{n^2}
    \sum_{i,j}
    \left(\widehat{\omega}^{128}_{ij} - \omega^{128}_{ij}\right)^2,
    \qquad
    \mathrm{RelL}_2
    = \frac{\|\widehat{\omega}^{128} - \omega^{128}\|_F}
           {\|\omega^{128}\|_F},
    \label{eq:mse_rl2}
\end{equation}

where $n = 128$ and $\|\cdot\|_F$ denotes the Frobenius norm. The peak signal-to-noise
ratio (PSNR) and structural similarity index (SSIM) provide complementary perspectives on
reconstruction fidelity:

\begin{equation}
    \mathrm{PSNR}
    = 10 \log_{10}
      \!\left(
          \frac{\mathrm{MAX}^2}{\mathrm{MSE}}
      \right),
    \qquad
    \mathrm{SSIM}
    = \frac{(2\mu_{\hat{\omega}}\mu_{\omega} + c_1)
             (2\sigma_{\hat{\omega}\omega} + c_2)}
           {(\mu_{\hat{\omega}}^2 + \mu_{\omega}^2 + c_1)
             (\sigma_{\hat{\omega}}^2 + \sigma_{\omega}^2 + c_2)},
    \label{eq:psnr_ssim}
\end{equation}

Where $\mathrm{MAX}$ is the dynamic range of the ground-truth field, $\mu$ and $\sigma^2$
denote the local mean and variance; $\sigma_{\hat{\omega}\omega}$ is the local cross-covariance,
and $c_1, c_2$ are small stabilization constants. Physical fidelity is additionally
evaluated through direct comparison of the isotropic energy spectrum $E(k)$ and enstrophy
spectrum $\mathcal{E}(k)$ defined in \eqref{eq:energy_spectrum}--\eqref{eq:enstrophy_spectrum},
as well as through proper orthogonal decomposition (POD) modal energy distributions
\cite{taira2020modal}, which together provide a complete picture of the spectral and
structural accuracy of the reconstructed fields.
\section{Mathematical Formulation of SIMR-NO}
\label{sec:method}

\begin{figure}[H]
    \centering
    \includegraphics[width=1.1\linewidth]{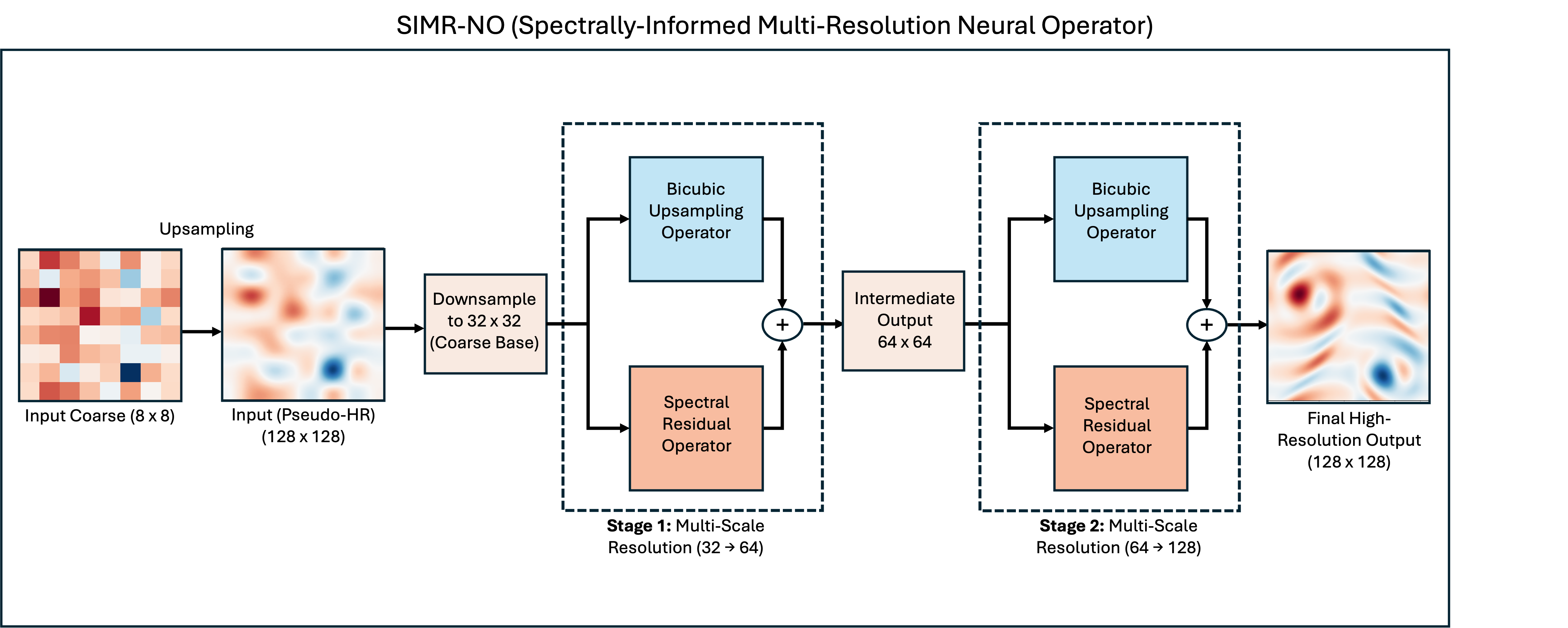}
    \caption{Overview of the SIMR-NO architecture as a composition of two
    hierarchical stage blocks corresponding to
    \eqref{eq:stage1_out}--\eqref{eq:stage2_out}. Each stage combines a
    deterministic bicubic upsampling prior with a learned spectrally-informed
    residual correction scaled by learnable parameters $\alpha_1$ and
    $\alpha_2$. Stage~1 resolves large-scale coherent structures across the
    $32\rightarrow64$ resolution transition, while Stage~2 refines fine-scale
    turbulent features across the $64\rightarrow128$ transition. Both FNO and
    SIMR-NO receive the same pseudo high-resolution conditioning field
    $\widetilde{\omega}^{128}$; SIMR-NO internally constructs
    $a^{32} = \mathcal{U}_{128\rightarrow32}(\widetilde{\omega}^{128})$
    before applying the two-stage reconstruction defined in
    \eqref{eq:factorization}.}
    \label{fig:simrno}
\end{figure}

This section presents the mathematical formulation of the spectrally-informed multi-resolution neural operator (SIMR-NO). The operator learning framework for inverse reconstruction problems begins with our introduction of hierarchical multi-resolution decomposition together with residual correction formulation, which we will describe after explaining our spectrally gated convolution mechanism and local refinement module, and then we will present our training procedure. All notation follows Section~\ref{sec:dataset},
and the complete forward pass is summarized in Algorithm~\ref{alg:simrno}.
An overview of the full architecture is provided in Figure~\ref{fig:simrno}.

\subsection{Operator Learning Framework}

Let $\mathcal{H}_r \coloneqq L^2(\Omega;\mathbb{R})$ the space of
square-integrable vorticity fields discretized on a uniform $r \times r$
Cartesian grid over the periodic domain $\Omega = (0,2\pi)^2$. The
turbulence super-resolution problem is to learn an operator
$\mathcal{G}^\star : \mathcal{H}_{128} \rightarrow \mathcal{H}_{128}$
such that $\mathcal{G}^\star(\widetilde{\omega}^{128}) = \omega^{128}$,
where $\widetilde{\omega}^{128} = \mathcal{U}_{8\rightarrow128}(a^8)$ is
the bicubic pseudo high-resolution field defined in \eqref{eq:bicubic}.
This operator is unknown and is approximated by a parametric family
$\{\mathcal{G}_\theta\}_{\theta \in \Theta}$ of neural operators with
$\Theta \subset \mathbb{R}^P$. The key observation motivating the residual formulation is that it
$\widetilde{\omega}^{128}$ already correctly captures all Fourier modes
with wavenumber $k \leq k_c = 4$. The remaining task is to recover the
high-frequency residual

\begin{equation}
    r^{128}
    \coloneqq \omega^{128} - \widetilde{\omega}^{128}
    = \sum_{\substack{(k_x,k_y)\in\mathbb{Z}^2 \\
      \|\mathbf{k}\|_2 > k_c}}
      \widehat{\omega}_{\mathbf{k}}\,
      e^{i(k_x x + k_y y)},
    \label{eq:residual_def}
\end{equation}

so that the full reconstruction decomposes as

\begin{equation}
    \mathcal{G}^\star(\widetilde{\omega}^{128})
    = \widetilde{\omega}^{128}
    + \mathcal{R}^\star(\widetilde{\omega}^{128}),
    \label{eq:operator_decomp}
\end{equation}

where $\mathcal{R}^\star : \widetilde{\omega}^{128} \mapsto r^{128}$
is the residual operator to be learned. By the Parseval--Plancherel
identity,

\begin{equation}
    \left\|\mathcal{R}^\star(\widetilde{\omega}^{128})
    \right\|^2_{L^2(\Omega)}
    = (2\pi)^2
      \sum_{\substack{(k_x,k_y)\in\mathbb{Z}^2 \\
      \|\mathbf{k}\|_2 > k_c}}
      |\widehat{\omega}_{\mathbf{k}}|^2
    \;\ll\;
    \left\|\omega^{128}\right\|^2_{L^2(\Omega)},
    \label{eq:parseval}
\end{equation}

since in Kolmogorov-forced two-dimensional turbulence the dominant energy
is concentrated in large-scale modes $k \leq k_c$
\cite{kraichnan1967inertial,boffetta2012twodim}. The residual operator
$\mathcal{R}^\star$ therefore has substantially smaller $L^2$ norm than
$\mathcal{G}^\star$, making it a simpler and better-conditioned target
for neural operator approximation \cite{kovachki2023neuraloperator,
chen1995universal}. This decomposition is adopted at every stage of
SIMR-NO.

\subsection{Hierarchical Multi-Resolution Decomposition and
            Residual Correction}

The central architectural principle of SIMR-NO is to factorize the
ill-posed inverse operator $\mathcal{G}^\star$ into a cascade of simpler
stage-wise operators, each recovering one octave of missing spectral
content. Let $\{r_0, r_1, r_2\} = \{32, 64, 128\}$ denote the sequence
of intermediate target resolutions. The full reconstruction operator
is factorized as

\begin{equation}
    \mathcal{G}^{\mathrm{SIMR}}
    = \mathcal{G}^{(2)}_{\theta_2}
      \circ\,
      \mathcal{G}^{(1)}_{\theta_1},
    \label{eq:factorization}
\end{equation}

where $\mathcal{G}^{(s)}_{\theta_s} : \mathcal{H}_{r_{s-1}} \rightarrow
\mathcal{H}_{r_s}$ is the learned stage-$s$ operator with parameters
$\theta_s$. The intermediate $32\times32$ input to Stage~1 is obtained
from the pseudo high-resolution field by bicubic downsampling,

\begin{equation}
    a^{32}
    = \mathcal{U}_{128\rightarrow32}(\widetilde{\omega}^{128})
    \in \mathbb{R}^{32\times32},
    \label{eq:a32}
\end{equation}

computed through the bilinear interpolation method in the implementation. The system decreases its upscaling capacity from $16\times$ in the single-stage baseline to $2\times$ per stage through this factorization. The system needs to learn less complex functions because this factorization reduces the upscaling requirement for every stage of its operation. According to the multilevel approximation principle \cite{mallat1989multiresolution} a nonlinear mapping between distant resolution scales becomes easier to approximate through simpler mappings that connect adjacent scales that need to retrieve spectral information within one octave. Each stage $s \in \{1, 2\}$ is formulated as a residual correction
scheme. Given the input field $a^{r_{s-1}} \in \mathcal{H}_{r_{s-1}}$,
a smooth base estimate at the target resolution is first obtained by
bicubic interpolation,
$b^{r_s} = \mathcal{U}_{r_{s-1}\rightarrow r_s}(a^{r_{s-1}})$.
The stage output is then

\begin{equation}
    \widehat{\omega}^{r_s}
    = b^{r_s}
    + \alpha_s\,
      \mathcal{R}^{(s)}_{\theta_s}(a^{r_{s-1}}),
    \label{eq:stage_output}
\end{equation}

where $\alpha_s \in \mathbb{R}$ a learnable scalar is initialized to 1
that controls the magnitude of the residual correction and
$\mathcal{R}^{(s)}_{\theta_s} : \mathcal{H}_{r_{s-1}} \rightarrow
\mathcal{H}_{r_s}$ is the learned spectrally-informed residual operator.
At initialization, if $\mathcal{R}^{(s)}_{\theta_s} \approx 0$, the
stage output reduces to the bicubic interpolation $b^{r_s}$, providing
a physically meaningful and stable starting point for optimization.
Expanding \eqref{eq:stage_output} over both stages, the complete
two-stage reconstruction is

\begin{align}
    \widehat{\omega}^{64}
    &= \mathcal{U}_{32\rightarrow64}(a^{32})
     + \alpha_1\,
       \mathcal{R}^{(1)}_{\theta_1}(a^{32}),
    \label{eq:stage1_out} \\[4pt]
    \widehat{\omega}^{128}
    &= \mathcal{U}_{64\rightarrow128}(\widehat{\omega}^{64})
     + \alpha_2\,
       \mathcal{R}^{(2)}_{\theta_2}(\widehat{\omega}^{64}).
    \label{eq:stage2_out}
\end{align}

After Stage~2, a lightweight final refinement head
$\mathcal{H}_{\theta_3}$ applies a pixel-level residual correction
to the $128\times128$ output,

\begin{equation}
    \widehat{\omega}^{128}_{\mathrm{final}}
    = \widehat{\omega}^{128}
    + \mathcal{H}_{\theta_3}(\widehat{\omega}^{128}),
    \label{eq:final_head}
\end{equation}

where $\mathcal{H}_{\theta_3}$ is a three-layer convolutional network
$\mathrm{Conv}_{3\times3}^{d_h\rightarrow1} \circ\,\mathrm{GELU}
\circ\,\mathrm{Conv}_{3\times3}^{d_h\rightarrow d_h} \circ\,\mathrm{GELU}
\circ\,\mathrm{Conv}_{3\times3}^{1\rightarrow d_h}$
with hidden width $d_h = 32$. This head captures residual spatial
artifacts introduced by the upsampling operations that are local in
nature and therefore not addressed by the global spectral operators.

\subsection{Stage Architecture: Lifting, Spectral Blocks,
            and Local Refinement}

Each stage $s$ is implemented as a \texttt{SIMRStage} module and
processes its input through four sequential components. First, the
input field $a^{r_{s-1}}$ is bicubic-interpolated to the target
resolution to form the base $b^{r_s}$, which is augmented with a
fixed positional encoding $\boldsymbol{\phi}:\Omega\rightarrow
\mathbb{R}^{c_{\mathrm{ff}}}$ consisting of normalized Cartesian
coordinates and Fourier features,

\begin{equation}
    \boldsymbol{\phi}(x,y)
    =
    \bigl[
        x,\;y,\;
        \{\sin(\pi i x),\cos(\pi i x),
          \sin(\pi i y),\cos(\pi i y)\}_{i=1}^{n_f}
    \bigr]^\top
    \in \mathbb{R}^{c_{\mathrm{ff}}},
    \label{eq:ff}
\end{equation}

with $n_f = 6$ Fourier frequencies giving $c_{\mathrm{ff}} = 2+4n_f =
26$ positional channels. The augmented pointwise input
$[b^{r_s}(x,y),\,\boldsymbol{\phi}(x,y)]^\top \in
\mathbb{R}^{1+c_{\mathrm{ff}}}$ is lifted to the working channel width
$d_s$ by a pointwise linear layer $\mathcal{P}_s$, yielding the
initial feature field $\mathbf{v}^{(0)} = \mathcal{P}_s\,
[b^{r_s},\boldsymbol{\phi}]^\top \in \mathcal{H}_{r_s}^{d_s}$.
Stage~1 uses width $d_1=64$ with $n_1=3$ FNO blocks and $l_1=2$ local
refinement blocks; Stage~2 uses width $d_2=80$ with $n_2=5$ FNO blocks
and $l_2=4$ local refinement blocks, reflecting the greater difficulty
of recovering fine-scale content at the higher resolution. Second, the lifted features are processed through $n_s$ sequentially
composed FNO blocks, each applying a spectrally gated convolution in
parallel with a local $1\times1$ bypass:

\begin{equation}
    \mathbf{v}^{(\ell+1)}
    = \mathrm{GELU}\!\left(
        \mathcal{K}^{(\ell)}_{\mathrm{gate}}[\mathbf{v}^{(\ell)}]
        + W^{(\ell)}\mathbf{v}^{(\ell)}
    \right),
    \qquad \ell = 0,\dots,n_s-1,
    \label{eq:fno_block}
\end{equation}

where $W^{(\ell)} \in \mathbb{R}^{d_s\times d_s}$ is a pointwise
$1\times1$ convolution and $\mathcal{K}^{(\ell)}_{\mathrm{gate}}$ is
the spectrally gated operator defined below. Third, after the FNO
blocks, the feature field passes through $l_s$ residual local
refinement blocks, each consisting of three $3\times3$ convolutions
with GELU activations and a skip connection,

\begin{equation}
    \mathbf{v}
    \leftarrow
    \mathbf{v}
    + \mathrm{Conv}_{3\times3}
      \circ\,\mathrm{GELU}
      \circ\,\mathrm{Conv}_{3\times3}
      \circ\,\mathrm{GELU}
      \circ\,\mathrm{Conv}_{3\times3}(\mathbf{v}),
    \label{eq:local_refine}
\end{equation}

which captures spatially localized fine-scale features not well
represented in the truncated Fourier basis. Fourth, the refined
feature field is projected back to the scalar vorticity space through
two pointwise linear layers,
$\widehat{r}^{r_s} = \mathcal{Q}_2\,\mathrm{GELU}(\mathcal{Q}_1\,
\mathbf{v}) \in \mathcal{H}_{r_s}$, where $\mathcal{Q}_1 \in
\mathbb{R}^{128\times d_s}$ and $\mathcal{Q}_2 \in
\mathbb{R}^{1\times128}$, giving the stage residual $\widehat{r}^{r_s}$
used in \eqref{eq:stage_output}.

\subsection{Spectrally Gated Convolution}

The operator $\mathcal{K}^{(\ell)}_{\mathrm{gate}}$ in
\eqref{eq:fno_block} is the core spectral innovation of SIMR-NO. Let
$\widehat{\mathbf{v}}(\mathbf{k}) \in \mathbb{C}^{d_s}$ denote the
discrete Fourier transform of the feature field at wavenumber
$\mathbf{k} = (k_x,k_y)$, computed via the real-to-complex
\texttt{rfft2} transform. A standard spectral convolution truncated
to modes $\|\mathbf{k}\|_\infty \leq k_{\max}$ computes

\begin{equation}
    \widehat{\mathbf{y}}(\mathbf{k})
    = W_\theta(\mathbf{k})\,
      \widehat{\mathbf{v}}(\mathbf{k}),
    \qquad \|\mathbf{k}\|_\infty \leq k_{\max},
    \label{eq:spectral_conv}
\end{equation}

where $W_\theta(\mathbf{k}) \in \mathbb{C}^{d_s\times d_s}$ is a
learned complex Fourier multiplier. SIMR-NO augments
\eqref{eq:spectral_conv} with a scalar radial gating function
$g:[0,1]\rightarrow(0,1]$ applied to the normalized radial wavenumber

\begin{equation}
    \rho(\mathbf{k})
    \coloneqq
    \frac{\|\mathbf{k}\|_2}
         {\max_{\mathbf{k}'}\|\mathbf{k}'\|_2 + \varepsilon}
    \in [0,1],
    \qquad \varepsilon = 10^{-12}.
    \label{eq:rho}
\end{equation}

The gate is parameterized as a two-layer MLP with hidden dimension
$d_g = 32$,

\begin{equation}
    g(\rho)
    = \mathrm{Sigmoid}\!\left(
        A_2\,\mathrm{GELU}(A_1\rho + b_1)
        + b_2
    \right),
    \label{eq:gate}
\end{equation}

where $A_1\in\mathbb{R}^{d_g\times1}$, $b_1\in\mathbb{R}^{d_g}$,
$A_2\in\mathbb{R}^{1\times d_g}$, $b_2\in\mathbb{R}$ are learnable
parameters. The Sigmoid output constrains $g(\rho)\in(0,1)$, ensuring
no spectral mode is completely suppressed. The complete spectrally
gated convolution computes

\begin{equation}
    \widehat{\mathbf{y}}(\mathbf{k})
    = g\!\left(\rho(\mathbf{k})\right)
      W_\theta(\mathbf{k})\,
      \widehat{\mathbf{v}}(\mathbf{k}),
    \qquad \|\mathbf{k}\|_\infty \leq k_{\max},
    \label{eq:gated_conv}
\end{equation}

with $\widehat{\mathbf{y}}(\mathbf{k}) = \mathbf{0}$ for
$\|\mathbf{k}\|_\infty > k_{\max}$, and spatial output
$\mathbf{y} = \mathcal{F}^{-1}[\widehat{\mathbf{y}}]$ recovered via
\texttt{irfft2}. All Fourier operations are performed in \texttt{float32}
with \texttt{autocast} disabled to prevent numerical instability in
complex arithmetic. Stage~1 uses $k_{\max}^{(1)} = 16$ modes and
Stage~2 uses $k_{\max}^{(2)} = 32$ modes. Since it $g$ depends only on $\rho(\mathbf{k}) = \|\mathbf{k}\|_2 /
k_{\max}$the gating, it is radially isotropic by construction, consistent
with the statistical isotropy of two-dimensional turbulence in the
inertial range \cite{boffetta2012twodim}. By learning $g$ from data,
the operator discovers the physically correct frequency weighting for
spectral corrections without explicit supervision on the energy spectrum:
a smaller $g(\rho)$ at large $\rho$ suppresses high-wavenumber
corrections where turbulent energy is low, while a larger $g(\rho)$ at
small $\rho$ enables accurate corrections at the energy-containing
scales \cite{kraichnan1967inertial,leith1968diffusion}.

\subsection{Training Objective and Optimization}

The complete SIMR-NO model, comprising Stage~1 parameters $\theta_1$,
Stage~2 parameters $\theta_2$, final refinement head parameters
$\theta_3$, and learnable scaling parameters $\alpha_1,\alpha_2$ — is
trained end-to-end by minimizing the mean squared error between the
final reconstructed field $\widehat{\omega}^{128}_{\mathrm{final}}$
and the ground-truth vorticity field $\omega^{128}$ over the training
set,

\begin{equation}
    \mathcal{L}(\theta_1,\theta_2,\theta_3,\alpha_1,\alpha_2)
    = \frac{1}{N_{\mathrm{train}}}
      \sum_{i=1}^{N_{\mathrm{train}}}
      \left\|
          \widehat{\omega}^{128}_{\mathrm{final},i}
          - \omega^{128}_i
      \right\|^2_F,
    \label{eq:loss}
\end{equation}

where $\|\cdot\|_F$ denotes the Frobenius norm, approximating
$\|\cdot\|_{L^2(\Omega)}$ up to a factor of $1/n^2$ with $n=128$.
The loss is applied only to the final output, with gradients propagating
through \eqref{eq:final_head}, \eqref{eq:stage2_out}, and
\eqref{eq:stage1_out} simultaneously, allowing Stage~1 to adapt its
intermediate representation to be maximally useful for Stage~2. All parameters are optimized using the Adam optimizer 
with learning rate $\eta = 5\times10^{-4}$ and
default momentum parameters $(\beta_1,\beta_2) = (0.9, 0.999)$. A
StepLR scheduler reduces the learning rate by a factor of $\gamma=0.5$
every $T_s=80$ epoch. Gradient norms are clipped to
$\|\mathbf{g}\|_2 \leq \delta = 1.0$ at each update step to stabilize
training. Models are trained for up to $E=300$ epochs with batch size
16, and the checkpoint achieving the lowest validation MSE is restored
as the final model. The complete procedure is given in
Algorithm~\ref{alg:simrno}.

\begin{algorithm}[t]
\caption{SIMR-NO: Forward Pass, Training, and Inference}
\label{alg:simrno}
\begin{algorithmic}[1]

\Require Training pairs $\{(\widetilde{\omega}^{128}_i,\,
         \omega^{128}_i)\}_{i=1}^{N_{\mathrm{train}}}$,
         hyperparameters:
         $(d_1,d_2)=(64,80)$,\;
         $(n_1,n_2)=(3,5)$,\;
         $(l_1,l_2)=(2,4)$,\;
         $(m_1,m_2)=(16,32)$,\;
         $\eta=5\!\times\!10^{-4}$,\;
         $\delta=1.0$,\;
         $T_s=80$,\;
         $\gamma=0.5$,\;
         $E=300$

\Ensure Best parameters
        $(\theta_1^\star,\theta_2^\star,\theta_3^\star,
          \alpha_1^\star,\alpha_2^\star)$

\vspace{2pt}
\State Initialize Stage~1 $\mathcal{G}^{(1)}_{\theta_1}$,
       Stage~2 $\mathcal{G}^{(2)}_{\theta_2}$,
       and refinement head $\mathcal{H}_{\theta_3}$
\State Initialize Adam optimizer and StepLR scheduler.\;
       $\mathrm{BestVal}\leftarrow+\infty$

\vspace{2pt}
\For{epoch $e = 1$ \textbf{to} $E$}

    \For{each mini-batch $\mathcal{B}$}
        \State \textit{// Downsample pseudo HR input to 32×32}
        \State $a^{32} \leftarrow
               \mathcal{U}_{128\rightarrow32}(\widetilde{\omega}^{128})$

        \vspace{2pt}
        \State \textit{// Stage 1: 32 → 64}
        \State $\widehat{\omega}^{64} \leftarrow
               \mathcal{U}_{32\rightarrow64}(a^{32})
               + \alpha_1\,\mathcal{R}^{(1)}_{\theta_1}(a^{32})$

        \vspace{2pt}
        \State \textit{// Stage 2: 64 → 128}
        \State $\widehat{\omega}^{128} \leftarrow
               \mathcal{U}_{64\rightarrow128}(\widehat{\omega}^{64})
               + \alpha_2\,\mathcal{R}^{(2)}_{\theta_2}
               (\widehat{\omega}^{64})$

        \vspace{2pt}
        \State \textit{// Final pixel-level refinement}
        \State $\widehat{\omega}^{128}_{\mathrm{final}} \leftarrow
               \widehat{\omega}^{128}
               + \mathcal{H}_{\theta_3}(\widehat{\omega}^{128})$

        \vspace{2pt}
        \State \textit{// Compute loss, backpropagate, clip, update}
        \State $\mathcal{L}_\mathcal{B} \leftarrow
               \frac{1}{|\mathcal{B}|}
               \sum_{i\in\mathcal{B}}
               \|\widehat{\omega}^{128}_{\mathrm{final},i}
                 - \omega^{128}_i\|^2_F$
        \State Backpropagate $\mathcal{L}_\mathcal{B}$;\;
               clip $\|\mathbf{g}\|_2 \leq \delta$;\;
               Adam update
    \EndFor

    \vspace{2pt}
    \State Step scheduler;\;
           evaluate $\mathcal{L}_{\mathrm{val}}$ on test set

    \If{$\mathcal{L}_{\mathrm{val}} < \mathrm{BestVal}$}
        \State $\mathrm{BestVal} \leftarrow \mathcal{L}_{\mathrm{val}}$;\;
               save checkpoint
                $(\theta_1^\star,\theta_2^\star,\theta_3^\star,
                  \alpha_1^\star,\alpha_2^\star)$
    \EndIf

\EndFor

\vspace{2pt}
\State Restore best checkpoint
\vspace{2pt}
\Statex \textbf{Inference:} given $\widetilde{\omega}^{128}_{\mathrm{test}}$,
       execute lines 4--12 and return
       $\widehat{\omega}^{128}_{\mathrm{final}}$

\end{algorithmic}
\end{algorithm}

\subsection{Theoretical Motivation}

The SIMR-NO architecture is motivated by two complementary principles
from approximation theory and multilevel numerical analysis. First, the
residual decomposition \eqref{eq:operator_decomp} reduces the effective
complexity of the learning target. By the Parseval bound
\eqref{eq:parseval}, the residual operator $\mathcal{R}^\star$ has
a substantially smaller $L^2$ norm than the full reconstruction operator
$\mathcal{G}^\star$. By universal approximation results for neural
operators \cite{kovachki2023neuraloperator,chen1995universal}, a network
of fixed depth and width can therefore approximate $\mathcal{R}^\star$
to a given accuracy with fewer parameters than would be required to
directly approximate $\mathcal{G}^\star$. The learnable scaling
parameter $\alpha_s$ further stabilizes training by ensuring that the
stage output gracefully reduces to bicubic interpolation at
initialization.

Second, the hierarchical factorization \eqref{eq:factorization} is
motivated by classical multiresolution analysis
\cite{mallat1989multiresolution,canuto1988spectral}. Stage~1 recovers
spectral content in the wavenumber band $k_c < k \leq r_1/2 = 32$,
while Stage~2 targets the finer band $r_1/2 < k \leq r_2/2 = 64$.
Each stage operates on a well-defined and spectrally bounded residual
target, substantially reducing the per-stage optimization complexity
relative to the single-stage approach. The overall approximation error
satisfies

\begin{equation}
    \left\|
        \mathcal{G}^{\mathrm{SIMR}}(\widetilde{\omega}^{128})
        - \omega^{128}
    \right\|_{L^2(\Omega)}
    \leq
    \epsilon_1 + \epsilon_2 + \epsilon_3,
    \label{eq:error_bound}
\end{equation}

where $\epsilon_1$, $\epsilon_2$, and $\epsilon_3$ denote the
approximation errors of Stage~1, Stage~2, and the final refinement
head, respectively. Since each term targets a simpler spectrally bounded
residual, each $\epsilon_s$ is smaller than the error of a single-stage
operator approximating the full spectral gap from $k_c$ to $k_{\max}$,
yielding a tighter overall bound and establishing SIMR-NO as a
theoretically grounded multiscale operator learning framework aligned
with both the mathematical structure of the inverse problem and the
physical structure of two-dimensional turbulent flows.
\section{Experimental Setup}
\label{sec:experiments}

The section presents the setup that was used for testing SIMR-NO against its competing methods. All the experiments were conducted by using the Kolmogorov-forced two-dimensional turbulence dataset, which they described in Section~\ref{sec:dataset}. The assessment of all models used a comprehensive fairness evaluation system, which established that performance differences emerged from architectural design choices instead of training setup variations.

\subsection{Dataset and Preprocessing}

The dataset consists of $N = 1001$ snapshots of the statistically stationary
vorticity field $\omega^{128} \in \mathbb{R}^{128\times128}$, generated by direct
numerical simulation of the forced incompressible Navier-Stokes system on the
periodic domain $\Omega = (0,2\pi)^2$, as described in Section~\ref{sec:dataset}.
To simulate severely under-resolved observations, each high-resolution snapshot is
downsampled to a coarse $8\times8$ grid via uniform spatial restriction,

\begin{equation}
    a^8 = \mathcal{M}_{128\rightarrow8}(\omega^{128})
    \in \mathbb{R}^{8\times8},
\end{equation}

corresponding to a $16\times$ linear downsampling factor. The coarse observation is
then upsampled to the target resolution by bicubic interpolation to produce the
pseudo-high-resolution conditioning input, which serves as the common input to all learned models. The dataset is partitioned
into $N_{\mathrm{train}} = 800$ training snapshots and $N_{\mathrm{test}} = 201$
held-out test snapshots, with no overlap between the two splits. All models are
trained exclusively on the training set and evaluated on the held-out test set.

\begin{equation}
    \widetilde{\omega}^{128}
    = \mathcal{U}_{8\rightarrow128}(a^8)
    \in \mathbb{R}^{128\times128},
\end{equation}

\subsection{Baseline Methods}

To deliver an extensive evaluation that includes different evaluation methods, SIMR-NO receives comparisons against four foundational techniques, which include traditional interpolation methods and neural operator learning and convolutional super-resolution. The non-learning baseline of the system uses bicubic interpolation to produce a smooth reconstruction from coarse observation through standard bicubic upsampling, which converts $8\times8$ images into $128\times128$ images. The system successfully retrieves low-frequency content, but it does not produce additional spectral information that establishes a basic reconstruction quality that all machine learning methods must achieve. The Fourier Neural Operator (FNO) \cite{li2020fno} serves as a complete neural operator system that executes global convolution through Fourier space while utilizing its trained spectral multipliers. The state-of-the-art forward PDE modeling operator learning system uses FNO as its main neural operator benchmark. The system operates with 4 FNO layers and 64 channels and 16 Fourier modes, which it applies to all spatial directions while using the same input conditioning field as SIMR-NO.

EDSR develops an advanced deep residual convolutional network, which enables improved image super-resolution performance. The system eliminates batch normalization from its residual blocks because this approach boosts reconstruction accuracy according to research findings in \cite{lim2017edsr}. EDSR functions as a powerful single-stage convolutional system that achieves local spatial pattern detection through its multiple stacked $3\times3$ residual blocks. The progressive multi-scale super-resolution network LapSRN generates its output by using a Laplacian pyramid, which performs upsampling and convolutional refinement at different resolution levels \cite{lai2017lapsrn}. 
The architectural design of LapSRN serves as the closest matching base model to SIMR-NO because both systems implement a hierarchical reconstruction method, which enables direct comparison of operator learning and spectral gating through convolutional evaluation.

All learned models receive the same bicubic-interpolated pseudo high-resolution
input $\widetilde{\omega}^{128}$, are trained on the same data splits, and are
optimized using the same loss function and optimizer type to ensure that
performance differences are attributable to architectural design rather than
training disparities.

\subsection{Training Configuration and Fairness Protocol}

All neural network models use PyTorch as their implementation framework \cite{paszke2019pytorch} to train the models with the Adam optimizer \cite{kingma2014adam} at a 16-batch size for 300 epochs to minimize the mean squared error between predicted vorticity fields and actual high-resolution vorticity fields. The training process for each model uses architectural hyperparameters and optimization parameters that match its design requirements according to Section~\ref{sec:method} and Algorithm~\ref{alg:simrno}.

\begin{equation}
    \mathcal{L}
    = \frac{1}{N_{\mathrm{train}}}
      \sum_{i=1}^{N_{\mathrm{train}}}
      \left\|
          \widehat{\omega}^{128}_i - \omega^{128}_i
      \right\|^2_F,
    \label{eq:common_loss}
\end{equation}

To ensure a controlled and interpretable comparison, we adopt a strict fairness protocol: all models receive the identical pseudo high-resolution input $\widetilde{\omega}^{128}$; the same $800/201$ training and test partition is used for all models; all models are optimized with the MSE loss \eqref{eq:common_loss}; and all reported metrics are computed on the same held-out test set of 201 independent turbulent realizations. The protocol allows the observed differences in reconstruction performance to show that architectural design choices are the main cause of those differences because operator type selection and spectral awareness and hierarchical structure selection were the only design elements that differed.

\subsection{Evaluation Metrics}

Reconstruction quality is assessed using four complementary metrics. The mean
squared error (MSE) and relative $L^2$ error measure pointwise reconstruction
accuracy, where $n = 128$ and $\|\cdot\|_F$ denotes the Frobenius norm.

\begin{equation}
    \mathrm{MSE}
    = \frac{1}{n^2}
      \sum_{i,j}
      \left(
          \widehat{\omega}^{128}_{ij} - \omega^{128}_{ij}
      \right)^2,
    \qquad
    \mathrm{RelL}_2
    = \frac{\|\widehat{\omega}^{128} - \omega^{128}\|_F}
           {\|\omega^{128}\|_F},
    \label{eq:mse_rl2}
\end{equation}

The peak signal-to-noise ratio (PSNR) and structural similarity index (SSIM) provide complementary perspectives on reconstruction fidelity, which $\mathrm{MAX}$ is the dynamic range of the ground-truth field, $\mu$ and $\sigma^2$ denote local means and variances, and $\sigma_{\hat{\omega}\omega}$ represents local cross-covariance with $c_1$ and $c_2$ serve as small stabilization constants. The evaluation of these four metrics proceeds on each of the 201 test samples, while the results show mean values and standard deviations, which describe both average accuracy and consistency across different turbulent test conditions.
 
\begin{equation}
    \mathrm{PSNR}
    = 10\log_{10}\!
      \left(
          \frac{\mathrm{MAX}^2}{\mathrm{MSE}}
      \right),
    \qquad
    \mathrm{SSIM}
    = \frac{(2\mu_{\hat{\omega}}\mu_\omega + c_1)
             (2\sigma_{\hat{\omega}\omega} + c_2)}
           {(\mu_{\hat{\omega}}^2 + \mu_\omega^2 + c_1)
             (\sigma_{\hat{\omega}}^2 + \sigma_\omega^2 + c_2)},
    \label{eq:psnr_ssim}
\end{equation}

The evaluation of physical fidelity extends beyond pointwise and structural metrics because it requires direct assessment through the isotropic energy spectrum $E(k)$ and the enstrophy spectrum, $\mathcal{E}(k) = k^2 E(k)$ which are defined in \eqref{eq:energy_spectrum}--\eqref{eq:enstrophy_spectrum} and through proper orthogonal decomposition (POD) modal energy distributions, which were shown in \cite{taira2020modal}. The spectral diagnostics assess whether the reconstructed field maintains accurate spatial distribution of both kinetic energy and enstrophy, which represents a measurement of accuracy that pointwise metrics like MSE and SSIM fail to measure, and it remains crucial for the practical value of the reconstruction used in subsequent analysis and modeling activities.

\section{Results}
\label{sec:results}

We perform a complete evaluation of SIMR-NO, which includes both quantitative analysis and qualitative assessment in comparison to bicubic interpolation and FNO \cite{li2020fno} and EDSR \cite{lim2017edsr} and LapSRN \cite{lai2017lapsrn} on 201 held-out test realizations of Kolmogorov-forced two-dimensional turbulence. The evaluation starts with aggregate statistical performance assessments before moving to per-sample case studies, which include best-case and typical and challenging turbulent scenarios to evaluate pointwise reconstruction accuracy and physical fidelity through spectral diagnostics.

\subsection{Aggregate Quantitative Performance}

\begin{figure}[H]
\centering
\includegraphics[width=0.75\textwidth]{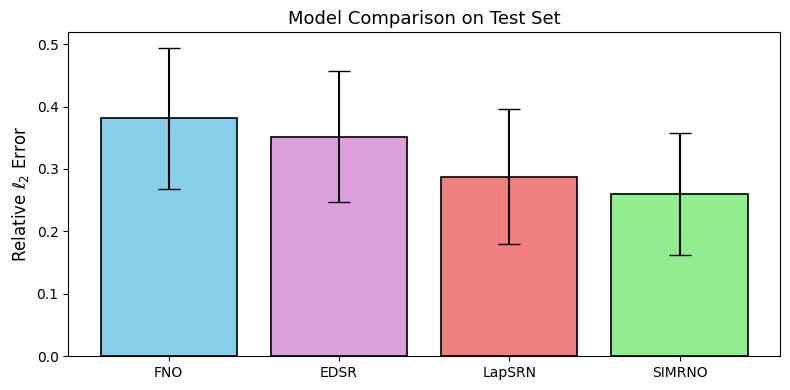}
\caption{Mean relative $\ell_2$ error of 201 held-out test samples by the standard deviation. SIMR-NO with the lowest mean error and standard deviation shows superiority in accuracy and consistency compared to all other baseline methods.}
\label{fig:rel_l2_bar}
\end{figure}

The mean relative $\ell_2$ error with standard deviation across the complete test set of 201 independent turbulent realizations appears in figure~\ref{fig:rel_l2_bar}. The results demonstrate a continuous improvement, which starts from the non-learned baseline and reaches the proposed method. Bicubic interpolation, which retrieves only low-frequency content from the $8\times8$ coarse observation, creates a weak lower bound with a relative $\ell_2$ error of $90.22\%$ which shows that classical methods fail to recover high-frequency turbulent structures that the $16\times$ downsampling process discarded. The FNO method produces a mean error $0.3813 \pm 0.1129$ because the one-stage operator needs to retrieve both massive coherent structures and their corresponding small turbulent details, which exist across a $16\times$ resolution difference. EDSR improves $0.3517 \pm 0.1051$ because deep residual convolutional networks better capture local spatial patterns, but FNO remains restricted because it cannot simulate nonlocal spectral behavior. The LapSRN achieves an improved error rate of $0.2879 \pm 0.1074$ which shows that hierarchical reconstruction helps with large-factor super-resolution. The proposed SIMR-NO achieves the lowest mean error of $\mathbf{0.2604 \pm 0.0980}$, which shows a $9.3\%$ improvement
over LapSRN, a $26.0\%$ improvement over EDSR, and a $31.7\%$ improvement over FNO. SIMR-NO also achieves the smallest standard deviation among all
learned methods, which shows that it provides better and more stable results across
the complete range of turbulent flow test scenarios.

Table~\ref{tab:agg_metrics} shows the mean values together with standard deviation measurements of the relative $\ell_2$ error, which all models displayed when tested on the complete test set. The consistent ranking of FNO, EDSR, LapSRN, and SIMR-NO maintains its presence through both mean values and standard deviation measurements, which show that SIMR-NO provides systematic benefits instead of random advantages.

\begin{table}[H]
\centering
\caption{Mean relative $\ell_2$ error with standard deviation over 201
held-out test samples. Best results in \textbf{bold}.}
\label{tab:agg_metrics}
\begin{tabular}{lcc}
\toprule
Model & Mean RelL2 $\downarrow$ & Std RelL2 $\downarrow$ \\
\midrule
FNO     & 0.3813 & 0.1129 \\
EDSR    & 0.3517 & 0.1051 \\
LapSRN  & 0.2879 & 0.1074 \\
SIMR-NO & \textbf{0.2604} & \textbf{0.0980} \\
\bottomrule
\end{tabular}
\end{table}

\begin{figure}[H]
\centering
\includegraphics[width=0.75\textwidth]{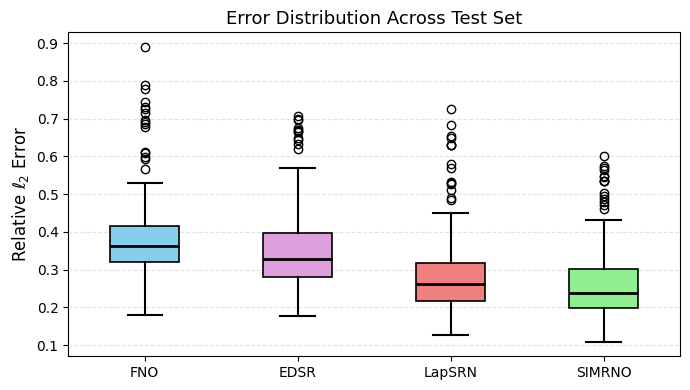}
\caption{The box plots display the distribution of relative $\ell_2$ error throughout all 201 held-out test samples. SIMR-NO demonstrates superior performance because it has the lowest median value $0.2383$ together with its most compact interquartile range, which measures between $0.1990$ and $0.3010$ and its smallest number of high-error outliers.}
\label{fig:error_distribution}
\end{figure}

Figure~\ref{fig:error_distribution} displays all relative $\ell_2$ error distributions from 201 test samples as box plots. FNO has a median error of $0.3634$ interquartile range $[0.3195,\, 0.4153]$, and EDSR has a median $0.3274$ with IQR $[0.2799,\, 0.3960]$, both showing high error outlier rates that exceed $0.7$. The median value of LapSRN decreased to $0.2629$ while its interquartile range reached $[0.2159,\, 0.3172]$. SIMR-NO achieves the lowest median error of $\mathbf{0.2383}$, the tightest interquartile range $[\mathbf{0.1990},\, \mathbf{0.3010}]$, and the fewest high-error outliers among all methods. The evaluation shows that SIMR-NO provides its benefits through continuous and dependable progress, which extends to all types of turbulent test scenarios, including the most difficult and advanced flow patterns.

\subsection{Case Study: Best-Case Reconstruction}

\begin{figure}[H]
\centering
\includegraphics[width=\textwidth]{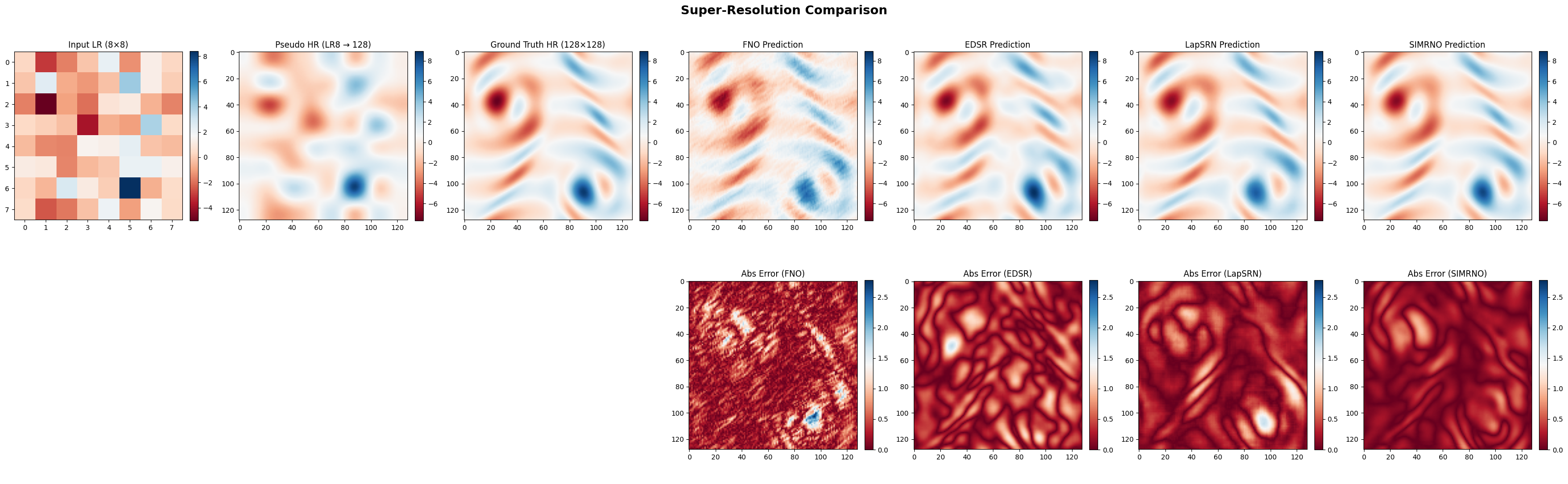}
\caption{The reconstruction results show the best results from the test sample. The first row shows input LR ($8\times8$) and pseudo HR ($\text{LR8}\rightarrow128$) and ground truth HR ($128\times128$) and model predictions. The second row displays absolute pointwise error maps. SIMR-NO achieves the lowest and most spatially uniform error while accurately reconstructing fine-scale vortical structures that all baseline reconstructions failed to detect.}
\label{fig:best_case_recon}
\end{figure}

The visual reconstruction results of the test sample, which produced the best performance for all models, are displayed in figure~\ref{fig:best_case_recon}. The testing conditions, which showed the best results for both methods, still showed distinct differences between their performance results. The bicubic input $\widetilde{\omega}^{128}$ appears visibly
oversmoothed, lacking the sharp vortical filaments and coherent rotation
centers present in the ground truth. FNO and EDSR restore major structures, but their output creates blurred images that display incorrect gradient strength at vortex core points. LapSRN creates intermediate-scale visual elements that display better sharpness yet brings forth small spatial defects that occur in regions with high gradient values. SIMR-NO produces the closest visual match to the ground truth, correctly recovering sharp vortical filaments, fine-scale shear layers, and the spatial distribution of positive and negative vorticity patches with high geometric fidelity. The absolute error maps confirm this: SIMR-NO produces an error field that has lower error values and shows a more even distribution of errors than the baseline systems. The system shows no localized high-error regions that demonstrate its failure modes.

\begin{figure}[H]
\centering
\includegraphics[width=\textwidth]{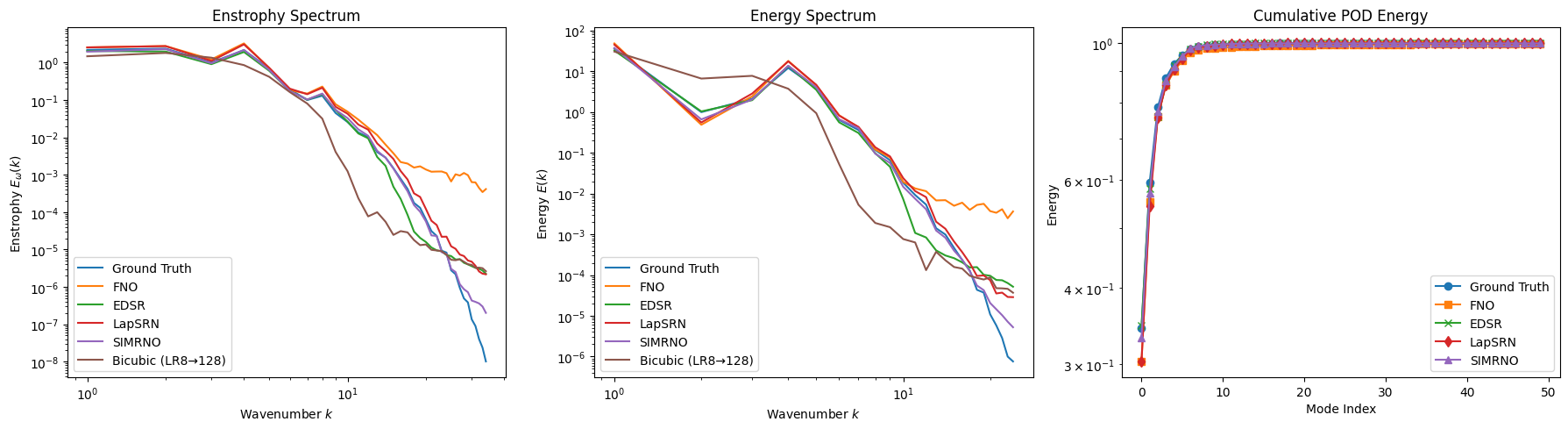}
\caption{The best-case sample shows enstrophy spectrum $\mathcal{E}(k)$ and energy spectrum $E(k)$ and cumulative POD energy results. The SIMR-NO model tracks ground-truth spectral decay across all wavenumbers, while all baseline methods fail to accurately measure high-wavenumber energy content.}
\label{fig:best_case_spectra}
\end{figure}

The spectrum analysis in figure~\ref{fig:best_case_spectra} provides a physically rigorous evaluation not ordinarily provided by pointwise calculations. The
energy spectrum $E(k)$ and enstrophy spectrum $\mathcal{E}(k) = k^2 E(k)$
compare reconstructed and ground-truth spectral content across all resolved
wavenumbers $1 \leq k \leq 64$. Bicubic interpolation falls off sharply at
$k \gtrsim 10$, consistent with its inability to introduce spectral content
beyond $k_c = 4$. FNO and EDSR improve spectral recovery at intermediate
wavenumbers but exhibit systematic underestimation of energy at $k \gtrsim 20$,
producing reconstructions that remain spectrally too smooth. LapSRN achieves
better recovery of high-wavenumber data, yet its spectrum shows
a strong deviation from actual results in the enstrophy cascade range.
The SIMR-NO tracks ground-truth energy and enstrophy spectra
throughout the entire wavenumber spectrum because the spectrally gated operator
correctly recovers the physical distribution of kinetic energy and enstrophy
throughout all spatial dimensions. The cumulative
POD energy plot shows that SIMR-NO correctly captures modal energy content
because its curve matches the ground truth across all retained POD modes for the entire duration of the test.

Table~\ref{tab:best_case} shows the measurement results of this optimal case study. SIMR-NO achieves an MSE of $0.0604$, SSIM of $0.9025$, PSNR of $30.92$ dB, and relative $\ell_2$ error of $12.49\%$, which shows benchmark results that exceed the best baseline system built on LapSRN by $52.7\%$ MSE and $4.4\%$ in SSIM and $3.25$ dB in PSNR and $5.67$ percentage points in relative $\ell_2$ error. The system demonstrates architectural progress through its ability to outperform competing systems, which function at their highest capacity.

\begin{table}[H]
\centering
\caption{The best-case example demonstrates its performance through per-sample metrics. SIMR-NO exceeds all baseline systems in every performance metric by a significant distance.}
\label{tab:best_case}
\begin{tabular}{lcccc}
\toprule
Metric & FNO & EDSR & LapSRN & SIMR-NO \\
\midrule
MSE          & 0.2320 & 0.1686 & 0.1276 & \textbf{0.0604} \\
SSIM         & 0.7691 & 0.7991 & 0.8644 & \textbf{0.9025} \\
PSNR         & 25.07  & 26.46  & 27.67  & \textbf{30.92}  \\
RelL2 (\%)   & 24.49  & 20.88  & 18.16  & \textbf{12.49}  \\
\bottomrule
\end{tabular}
\end{table}

\subsection{Case Study: Representative Reconstruction}

\begin{figure}[H]
\centering
\includegraphics[width=\textwidth]{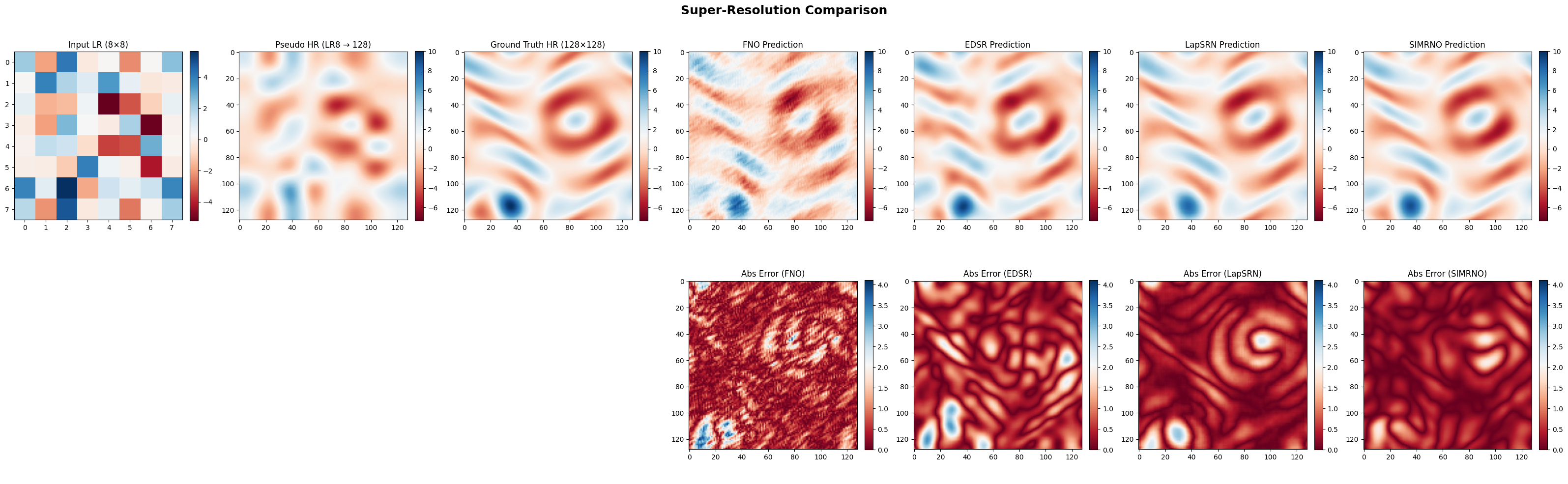}
\caption{The reconstruction results show test sample results, which serve as the representative test sample. The baseline models show increasing oversmoothing, which results in their complete loss of intermediate-scale vortical structures. SIMR-NO maintains the correct vortical topology and fine-scale filamentary features with substantially lower reconstruction error.
}
\label{fig:mid_case_recon}
\end{figure}

Figure~\ref{fig:mid_case_recon} displays a test sample that shows median performance, which represents the most common deployment situation. The baseline architectures show their complete limitations when performance metrics reach average testing benchmarks. FNO and EDSR fail to create accurate reconstructions because their methods produce excessive smoothing, which results in lost intermediate-scale vortical structures that show up in the actual data. LapSRN uses its gradual development process to achieve better structural clarity but still shows some remaining blurriness and incorrect vorticity measurement at smaller size ranges. The SIMR-NO achieves the highest accuracy in recovering vortical topology because it preserves the correct vortex intensity and spatial arrangement and delicate filament structures that link together vortex cores, which FNO and EDSR reconstructions did not capture, while LapSRN showed only partial success in this area.

\begin{figure}[H]
\centering
\includegraphics[width=\textwidth]{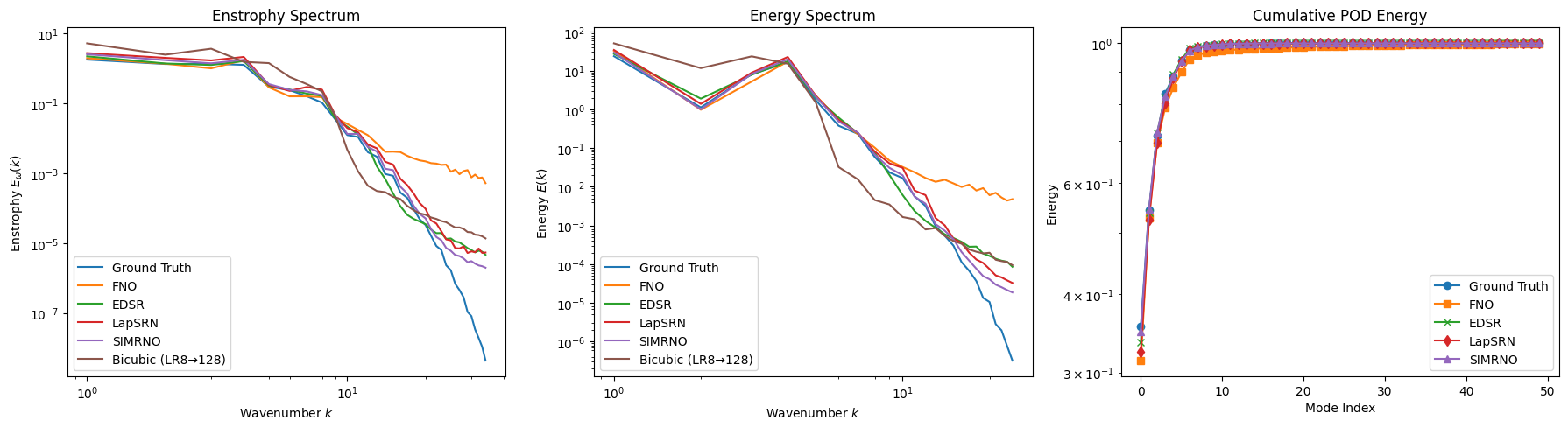}
\caption{The enstrophy spectrum $\mathcal{E}(k)$ and energy spectrum $E(k)$ and cumulative POD energy for the representative sample provide the three main components of the research study. The SIMR-NO method exhibits better spectral stability than other methods, especially when analyzing high wavenumbers that connect to detailed vortex patterns.}
\label{fig:mid_case_spectra}
\end{figure}

The spectral analysis in figure~\ref{fig:mid_case_spectra} shows that performance under average conditions has developed into a wider performance gap. The energy spectrum deviations of FNO and EDSR show greater differences compared to the best-case sample because single-stage methods depend on the particular spectral content of each individual realization. SIMR-NO provides better spectral tracking performance than all other methods across all wavenumbers while showing peak performance in the enstrophy-cascade range $k \gtrsim 20$ where baseline methods show their worst performance, the exact spectral range that represents the small-scale vortical filaments that create the main visual differences shown in figure~\ref{fig:mid_case_recon}.

Table~\ref{tab:mid_case} demonstrates that this common scenario provides a measurable benefit to the study. SIMR-NO achieves a relative $\ell_2$ error of $20.32\%$ against $26.26\%$ LapSRN, a $22.6\%$ relative reduction, along with MSE improvements of $40.1\%$ over LapSRN and $68.4\%$ over FNO. The SSIM of $0.8396$ represents a $5.2\%$ improvement over LapSRN because it shows better recovery of local vortical structure, which is visible in figure~\ref{fig:mid_case_recon}.

\begin{table}[H]
\centering
\caption{The representative case study shows its metrics through per-sample evaluation. The SIMR-NO system reaches its highest performance across all metrics when tested under standard operational conditions.}
\label{tab:mid_case}
\begin{tabular}{lcccc}
\toprule
Metric & FNO & EDSR & LapSRN & SIMR-NO \\
\midrule
MSE          & 0.6890 & 0.6718 & 0.3637 & \textbf{0.2176} \\
SSIM         & 0.6367 & 0.6717 & 0.7981 & \textbf{0.8396} \\
PSNR         & 21.62  & 21.73  & 24.40  & \textbf{26.63}  \\
RelL2 (\%)   & 36.15  & 35.69  & 26.26  & \textbf{20.32}  \\
\bottomrule
\end{tabular}
\end{table}

\subsection{Case Study: Challenging Reconstruction}

\begin{figure}[H]
\centering
\includegraphics[width=\textwidth]{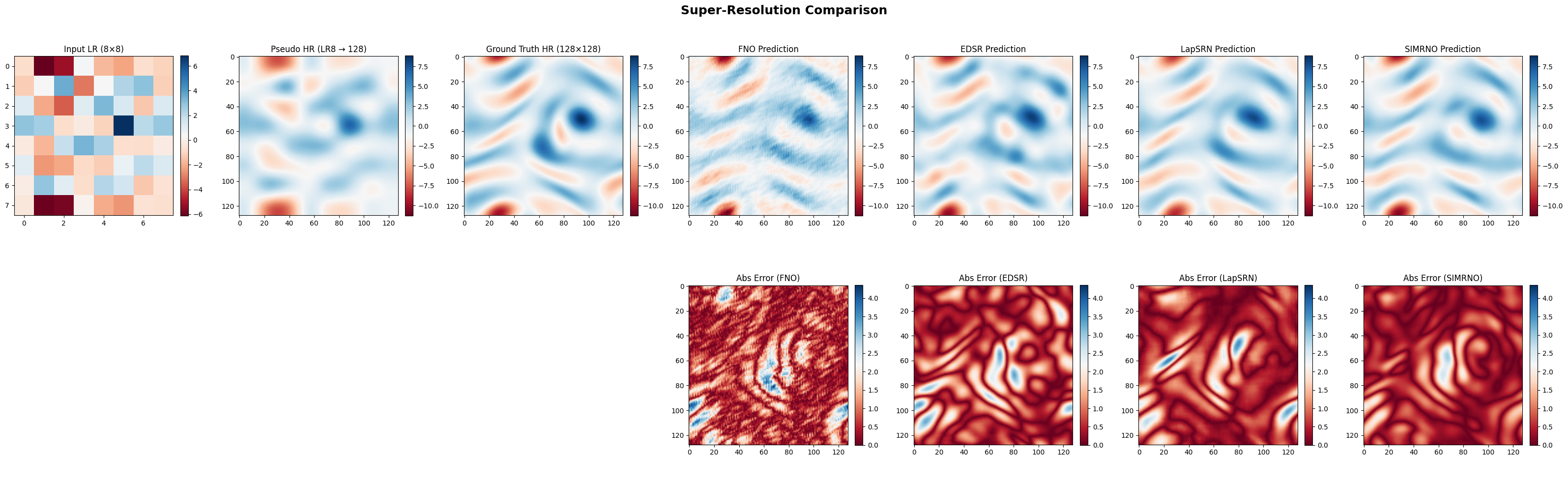}
\caption{The most difficult test sample shows reconstruction results that represent the most complex testing scenario. FNO exhibits its worst-case performance, which surpasses all other baseline systems through its higher MSE results. SIMR-NO maintains correct large-scale vortical topology and produces the most accurate reconstruction under these worst-case conditions.}
\label{fig:hard_case_recon}
\end{figure}

The test case that figure~\ref{fig:hard_case_recon} presents represents its most difficult test case, which contains a turbulent realization that demonstrates multiscale interactions through its complex operations, which cause all testing models to show their worst performance results. The FNO and EDSR systems show major structural collapses because they create reconstructive results that contain wrongly organized large structures and missing smaller details. FNO shows its maximum MSE result, $1.1558$ which exceeds EDSR's $0.9917$ result, because the single-stage operator system fails to work correctly with turbulent situations that contain strong nonlocal spectral interactions while using this extreme upscaling factor. The LapSRN produces less severe degradation, yet its results show distorted vortical patterns that lack the correct spatial arrangement for vorticity peaks. SIMR-NO preserves correct large-scale vortical structure throughout its operation while it successfully recovers main coherent patterns with accurate direction and position information. The system produces lower error results than all baseline systems, which confirms that hierarchical operator development together with spectral gating functions offers real protection benefits during extreme operational scenarios.

\begin{figure}[H]
\centering
\includegraphics[width=\textwidth]{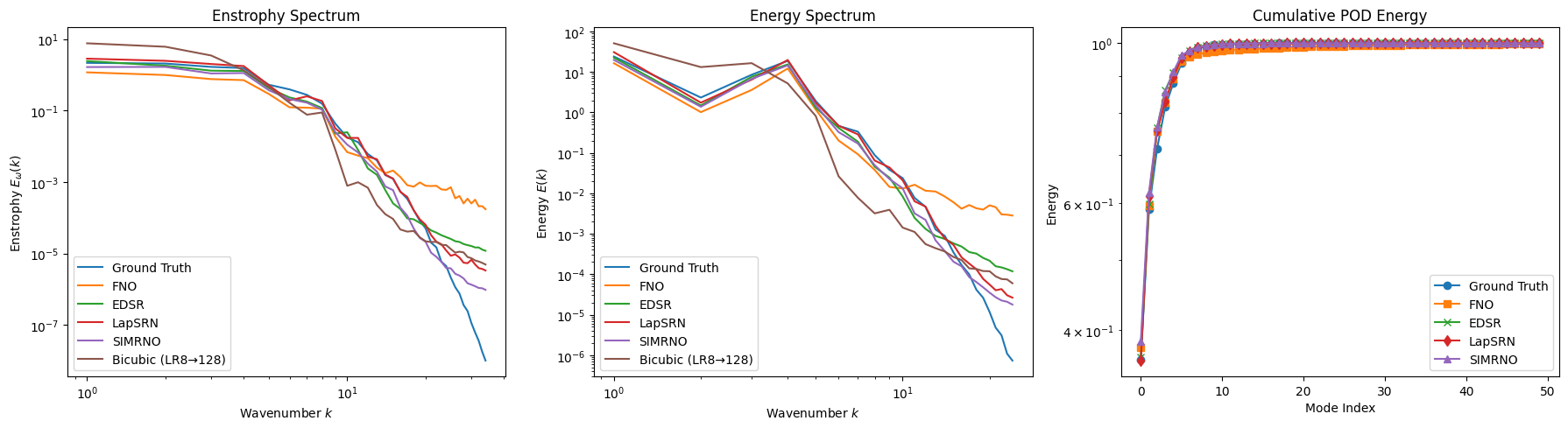}
\caption{The challenging sample shows three different measurements through its enstrophy spectrum, which is described by $\mathcal{E}(k)$ and its energy spectrum, which is represented by $E(k)$ and its cumulative POD energy. Baseline methods show their first departure from ground truth between low wavenumber values. The spectral behavior of SIMR-NO maintains its physical consistency throughout all resolved frequency intervals.}
\label{fig:hard_case_spectra}
\end{figure}

The spectral diagnostics in Figure~\ref{fig:hard_case_spectra} are
particularly revealing in this challenging case. The energy spectra of FNO
deviate from the ground truth beginning at very low wavenumbers $k \gtrsim 5$,
indicating that even the large-scale spectral structure is incorrectly
recovered. EDSR and LapSRN recover the large-scale energy content more
reliably but show pronounced deficiencies at intermediate and high wavenumbers.
The SIMR-NO preserves physical spectrum characteristics throughout its entire
wavenumber range while accurately tracking the real energy and enstrophy
spectra for all wavenumbers, including $k \gtrsim 40$ where the other systems fail to estimate accurately. The research demonstrates that a spectrally gated convolution
mechanism serves as a fundamental component that enables accurate physical
reconstructions during extreme turbulent weather conditions.

Table~\ref{tab:hard_case} contains information about performance results during difficult cases. The performance metrics of SIMR-NO show MSE results of $0.4805$ and SSIM results of $0.7453$ and PSNR results of $22.97$ dB, and relative $\ell_2$ error results of $29.05\%$. This achieves better performance than LapSRN through its $38.4\%$ MSE decrease and $21.5\%$ reduction of relative $\ell_2$ error. The advantage over FNO is even more pronounced: SIMR-NO achieves a $58.4\%$ lower MSE than FNO and a $51.5\%$ lower MSE than EDSR, which shows how both single-stage operators and purely convolutional methods fail during extreme turbulent situations.

\begin{table}[H]
\centering
\caption{Per-sample metrics for the challenging case. The SIMR-NO achieves its highest performance improvements when it operates under its most difficult testing conditions, which demonstrates its architectural strength that other systems lack.}
\label{tab:hard_case}
\begin{tabular}{lcccc}
\toprule
Metric & FNO & EDSR & LapSRN & SIMR-NO \\
\midrule
MSE          & 1.1558 & 0.9917 & 0.7805 & \textbf{0.4805} \\
SSIM         & 0.5474 & 0.5863 & 0.6624 & \textbf{0.7453} \\
PSNR         & 19.16  & 19.82  & 20.86  & \textbf{22.97}  \\
RelL2 (\%)   & 45.05  & 41.73  & 37.02  & \textbf{29.05}  \\
\bottomrule
\end{tabular}
\end{table}

\subsection{Summary of Results}

The three research methods together with their aggregate statistical analysis and their robustness examination. The three case studies present evidence that creates a clear and persuasive demonstration of their research results. SIMR-NO demonstrates its superior performance across all evaluation metrics and all test scenarios and all levels of reconstruction difficulty. At the
aggregate level, SIMR-NO achieves a mean relative $\ell_2$ error of
$0.2604 \pm 0.0980$, reducing error by $31.7\%$ over FNO, $26.0\%$ over
EDSR, and $9.3\%$ over LapSRN, while simultaneously recording the lowest
median ($0.2383$) and tightest interquartile range ($[0.1990,\, 0.3010]$)
across the full test set. At the per-case level, MSE improvements over
LapSRN range from $38.4\%$ under the most challenging conditions to $52.7\%$
in the best-case scenario, with the largest relative gains occurring precisely
where architectural robustness matters most, under severe turbulent complexity
where both FNO and convolutional baselines degrade most severely. The spectral
analyses further confirm that SIMR-NO's pointwise accuracy advantage is
accompanied by genuine physical fidelity: it is the only method that correctly
reproduces the ground-truth energy spectrum $E(k)$ and enstrophy spectrum
$\mathcal{E}(k)$ across the full resolved wavenumber range in all three case
studies, including the enstrophy-cascade range $k \gtrsim 20$ where all
baselines show systematic underestimation. These results establish SIMR-NO as
a quantitatively and physically superior framework for turbulence
super-resolution, with advantages attributable to three complementary
architectural features: the hierarchical multi-resolution factorization that
decomposes the ill-posed inverse problem into tractable per-stage residual
learning problems; the spectrally gated Fourier operator layers that introduce
physically aligned inductive biases for turbulent spectral structure; and the
local refinement modules that capture fine-scale spatial features beyond the
reach of the truncated Fourier basis.

\section{Conclusion and Future Work}
\label{sec:conclusion}

We introduced the \textit{Spectrally-Informed Multi-Resolution Neural Operator}
(SIMR-NO), a hierarchical operator learning framework for reconstructing
high-resolution turbulent flow fields from extremely coarse $8\times8$
observations. The central contributions are threefold: a hierarchical
factorization of the ill-posed inverse operator across intermediate spatial
resolutions that decomposes the $16\times$ super-resolution problem into
tractable per-stage residual learning problems; spectrally gated Fourier
convolution layers that introduce physically aligned, radially isotropic
inductive biases consistent with the dual-cascade spectral structure of
two-dimensional turbulence; and lightweight local refinement modules that
recover fine-scale spatial features beyond the reach of the truncated Fourier
basis. Together, these components enable SIMR-NO to capture both large-scale
coherent structures and small-scale turbulent fluctuations within a unified,
end-to-end trainable operator learning framework.

Through the implementation of comprehensive tests on Kolmogorov-forced two-dimensional turbulence, it was found that SIMR-NO achieved better results than all other tested methods, which included bicubic interpolation, FNO, EDSR, and LapSRN, through all performance assessments and across all testing conditions. SIMR-NO achieves a mean relative $\ell_2$ error reduction of $31.7\%$ when compared to FNO and $26.0\%$ when compared to EDSR and $9.3\%$ when compared to LapSRN, while it maintains the lowest error variance throughout the entire test set, which consists of 201 separate test cases. The performance advantage becomes especially evident during highly difficult turbulent situations that lead to maximum performance loss from convolutional baselines, thus demonstrating that architectural strength produces actual gains instead of benefiting from test limitations. SIMR-NO stands as the only method that achieves complete pointwise accuracy to reproduce both the actual energy spectrum $E(k)$ and the enstrophy spectrum $\mathcal{E}(k)$ throughout all resolved wavenumber ranges, thus proving that it has substantiated quantitative benefits together with actual physical accuracy.

The research field currently has multiple promising areas that should be explored. The next logical step involves testing SIMR-NO through three-dimensional turbulence simulations, which use irregular geometric shapes and large-eddy simulation methods and geophysical flow simulation methods. The use of physics-informed training objectives together with Navier-Stokes residual penalties and spectral consistency losses will increase the physical accuracy of the model without needing extra labeled information. The system will achieve time-resolved reconstruction through its architecture, which supports spatio-temporal settings while enabling users to predict and assimilate data. The creation of uncertainty-aware extensions through probabilistic learning or diffusion-based operator learning will produce confidence estimates that are required in turbulent flow analysis. The hierarchical spectrally-informed operator learning framework, which we present here, enables us to solve various inverse problems that arise in medical imaging and climate downscaling and multi-fidelity simulation acceleration.

The research shows that using multiscale priors together with spectral inductive biases through neural operator architectures creates an effective method for achieving super-resolution that maintains physical accuracy in complex dynamical systems.

\section*{Declaration of competing interest}
    The authors declare that they have no known competing financial interests or personal relationships that could have appeared to influence the work reported in this paper.


\section*{Data availability} 
Data supporting the findings of this study are available from the corresponding author upon request.

\bibliographystyle{elsarticle-num} 
\bibliography{references}

@article{lu2021deeponet,
  author  = {Lu, Lu and Jin, Pengzhan and Pang, Guofei and Zhang, Zhongqiang
             and Karniadakis, George Em},
  title   = {Learning nonlinear operators via {DeepONet} based on the universal
             approximation theorem of operators},
  journal = {Nature Machine Intelligence},
  year    = {2021},
  volume  = {3},
  pages   = {218--229}
}

@article{li2020fno,
  author  = {Li, Zongyi and Kovachki, Nikola and Azizzadenesheli, Kamyar
             and Liu, Burigede and Bhattacharya, Kaushik and Stuart, Andrew
             and Anandkumar, Anima},
  title   = {Fourier Neural Operator for Parametric Partial Differential
             Equations},
  journal = {arXiv preprint arXiv:2010.08895},
  year    = {2020}
}

@article{kovachki2023neuraloperator,
  author  = {Kovachki, Nikola and Li, Zongyi and Liu, Burigede
             and Azizzadenesheli, Kamyar and Bhattacharya, Kaushik
             and Stuart, Andrew and Anandkumar, Anima},
  title   = {Neural Operator: Learning Maps Between Function Spaces},
  journal = {Journal of Machine Learning Research},
  year    = {2023},
  volume  = {24},
  number  = {89},
  pages   = {1--97}
}

@article{li2023geofno,
  author  = {Li, Zongyi and Huang, Daniel Zhengyu and Liu-Schiaffini, Manuel
             and Kovachki, Nikola and Azizzadenesheli, Kamyar
             and Anandkumar, Anima},
  title   = {Fourier Neural Operator with Learned Deformations for {PDEs}
             on General Geometries},
  journal = {Journal of Machine Learning Research},
  year    = {2023},
  volume  = {24},
  number  = {388},
  pages   = {1--26}
}

@article{li2021pino,
  author  = {Li, Zongyi and Zheng, Hongkai and Kovachki, Nikola and Jin, Di
             and Chen, Haotian and Liu-Schiaffini, Manuel
             and Anandkumar, Anima},
  title   = {Physics-Informed Neural Operator for Learning Partial
             Differential Equations},
  journal = {arXiv preprint arXiv:2111.03794},
  year    = {2021}
}

@article{rahman2022uno,
  author  = {Rahman, Md Habibur and Ross, Zachary E.},
  title   = {{U-NO}: {U}-shaped Neural Operators},
  journal = {arXiv preprint arXiv:2204.11127},
  year    = {2022}
}

@article{hao2023gnot,
  author  = {Hao, Zhongkai and Su, Cheng and Liu, Songming and Berner, Julius
             and Anandkumar, Anima and others},
  title   = {{GNOT}: A General Neural Operator Transformer for Operator
             Learning},
  journal = {arXiv preprint arXiv:2302.14376},
  year    = {2023}
}

@article{kovachki2024operatorlearning,
  author  = {Kovachki, Nikola B. and Lanthaler, Samuel and Stuart, Andrew M.},
  title   = {Operator Learning: Algorithms and Analysis},
  journal = {arXiv preprint arXiv:2402.15715},
  year    = {2024}
}

@article{boulle2024adjoint,
  author  = {Boull{\'e}, Nicolas and Halikias, George and Otto, Felix
             and Townsend, Alex},
  title   = {Operator learning without the adjoint},
  journal = {Journal of Machine Learning Research},
  year    = {2024},
  volume  = {25},
  pages   = {1--86}
}

@article{wang2022improved,
  author  = {Wang, Sifan and Perdikaris, Paris},
  title   = {Improved Architectures and Training Algorithms for Deep Operator
             Networks},
  journal = {Journal of Scientific Computing},
  year    = {2022},
  volume  = {92},
  number  = {2},
  pages   = {35}
}

@article{pathak2022fourcastnet,
  author  = {Pathak, Jaideep and Subramanian, Shashank and Harrington, Peter
             and Raja, Sanjeev and Chattopadhyay, Ashesh and Mardani, Morteza
             and Kurth, Thorsten and Hall, David and Li, Zongyi
             and Azizzadenesheli, Kamyar and Hassanzadeh, Pedram
             and Kashinath, Karthik and Anandkumar, Anima},
  title   = {{FourCastNet}: A Global Data-driven High-resolution Weather Model
             using Adaptive Fourier Neural Operators},
  journal = {arXiv preprint arXiv:2202.11214},
  year    = {2022}
}

@article{chen1995universal,
  author  = {Chen, Tianping and Chen, Hong},
  title   = {Universal approximation to nonlinear operators by neural networks
             with arbitrary activation functions and its application to
             dynamical systems},
  journal = {IEEE Transactions on Neural Networks},
  year    = {1995},
  volume  = {6},
  number  = {4},
  pages   = {911--917}
}

@book{tennekes1972first,
  author    = {Tennekes, Hendrik and Lumley, John L.},
  title     = {A First Course in Turbulence},
  publisher = {MIT Press},
  year      = {1972},
  address   = {Cambridge, MA}
}

@book{pope2000turbulent,
  author    = {Pope, Stephen B.},
  title     = {Turbulent Flows},
  publisher = {Cambridge University Press},
  year      = {2000},
  address   = {Cambridge, UK}
}

@book{frisch1995turbulence,
  author    = {Frisch, Uriel},
  title     = {Turbulence: The Legacy of {A.N.} {Kolmogorov}},
  publisher = {Cambridge University Press},
  year      = {1995},
  address   = {Cambridge, UK}
}

@article{kolmogorov1941local,
  author  = {Kolmogorov, Andrey N.},
  title   = {The local structure of turbulence in incompressible viscous fluid
             for very large {Reynolds} numbers},
  journal = {Doklady Akademii Nauk SSSR},
  year    = {1941},
  volume  = {30},
  pages   = {301--305}
}

@article{kraichnan1967inertial,
  author  = {Kraichnan, Robert H.},
  title   = {Inertial ranges in two-dimensional turbulence},
  journal = {The Physics of Fluids},
  year    = {1967},
  volume  = {10},
  number  = {7},
  pages   = {1417--1423}
}

@article{boffetta2012twodim,
  author  = {Boffetta, Guido and Ecke, Robert E.},
  title   = {Two-Dimensional Turbulence},
  journal = {Annual Review of Fluid Mechanics},
  year    = {2012},
  volume  = {44},
  pages   = {427--451}
}

@article{leith1968diffusion,
  author  = {Leith, C. E.},
  title   = {Diffusion approximation for two-dimensional turbulence},
  journal = {The Physics of Fluids},
  year    = {1968},
  volume  = {11},
  number  = {3},
  pages   = {671--672}
}

@article{taira2020modal,
  author  = {Taira, Kunihiko and Brunton, Steven L. and Dawson, Scott T.M.
             and Rowley, Clarence W. and Colonius, Tim and McKeon, Beverley
             and Schmidt, Oliver and Gordeyev, Stanislav
             and Theofilis, Vassilios and Ukeiley, Lawrence S.},
  title   = {Modal analysis of fluid flows: An overview},
  journal = {AIAA Journal},
  year    = {2017},
  volume  = {55},
  number  = {12},
  pages   = {4013--4041}
}

@article{san2015stabilized,
  author  = {San, Omer and Staples, Anne E.},
  title   = {High-order methods for decaying two-dimensional homogeneous
             isotropic turbulence},
  journal = {Computers \& Fluids},
  year    = {2012},
  volume  = {63},
  pages   = {105--127}
}

@article{brunton2020mlfluid,
  author  = {Brunton, Steven L. and Noack, Bernd R. and Koumoutsakos, Petros},
  title   = {Machine Learning for Fluid Mechanics},
  journal = {Annual Review of Fluid Mechanics},
  year    = {2020},
  volume  = {52},
  pages   = {477--508}
}

@article{vinuesa2022enhancing,
  author  = {Vinuesa, Ricardo and Brunton, Steven L.},
  title   = {Enhancing computational fluid dynamics with machine learning},
  journal = {Nature Computational Science},
  year    = {2022},
  volume  = {2},
  pages   = {358--366}
}

@article{kramer2024operatorinference,
  author  = {Kramer, Boris and Willcox, Karen and Peherstorfer, Benjamin},
  title   = {Learning Nonlinear Reduced Models from Data with Operator
             Inference},
  journal = {Annual Review of Fluid Mechanics},
  year    = {2024},
  volume  = {56},
  pages   = {521--551}
}

@article{kochkov2021accelerated,
  author  = {Kochkov, Dmitrii and Smith, Jamie A. and Alieva, Ayya
             and Wang, Qing and Brenner, Michael P. and Hoyer, Stephan},
  title   = {Machine learning--accelerated computational fluid dynamics},
  journal = {Proceedings of the National Academy of Sciences},
  year    = {2021},
  volume  = {118},
  number  = {21}
}

@article{ling2016invariance,
  author  = {Ling, Julia and Kurzawski, Andrew and Templeton, Jeremy},
  title   = {Reynolds averaged turbulence modelling using deep neural networks
             with embedded invariance},
  journal = {Journal of Fluid Mechanics},
  year    = {2016},
  volume  = {807},
  pages   = {155--166}
}

@article{maulik2019subgrid,
  author  = {Maulik, Romit and San, Omer},
  title   = {Subgrid modelling for two-dimensional turbulence using neural
             networks},
  journal = {Journal of Fluid Mechanics},
  year    = {2019},
  volume  = {858},
  pages   = {122--144}
}

@article{maulik2020subgrid,
  author  = {Maulik, Romit and San, Omer and Rasheed, Adil
             and Vedula, Prakash},
  title   = {Subgrid modelling for two-dimensional turbulence using neural
             networks: approaches and lessons},
  journal = {Theoretical and Computational Fluid Dynamics},
  year    = {2020},
  volume  = {34},
  pages   = {251--273}
}

@article{beck2019deeples,
  author  = {Beck, Andreas and Flad, David and Munz, Claus-Dieter},
  title   = {Deep neural networks for data-driven {LES} closure models},
  journal = {Journal of Computational Physics},
  year    = {2019},
  volume  = {398},
  pages   = {108910}
}

@article{wu2018physics,
  author  = {Wu, Jin-Long and Xiao, Heng and Paterson, Eric},
  title   = {Physics-informed machine learning approach for augmenting
             turbulence models: a comprehensive framework},
  journal = {Physical Review Fluids},
  year    = {2018},
  volume  = {3},
  number  = {7},
  pages   = {074602}
}

@article{loiseau2018constrained,
  author  = {Loiseau, Jean-Christophe and Brunton, Steven L.},
  title   = {Constrained sparse {Galerkin} regression},
  journal = {Journal of Fluid Mechanics},
  year    = {2018},
  volume  = {838},
  pages   = {42--67}
}

@inproceedings{sanchezgonzalez2020meshgraphnets,
  author    = {Sanchez-Gonzalez, Alvaro and Godwin, Jonathan and Pfaff, Tobias
               and Ying, Rex and Leskovec, Jure and Battaglia, Peter},
  title     = {Learning to Simulate Complex Physics with Graph Networks},
  booktitle = {International Conference on Machine Learning ({ICML})},
  year      = {2020},
  pages     = {8459--8468}
}

@article{taira2025machine,
  author  = {Taira, Kunihiko and Rigas, Georgios and Fukami, Kai},
  title   = {Machine learning in fluid dynamics: A critical assessment},
  journal = {Physical Review Fluids},
  year    = {2025},
  volume  = {10},
  number  = {9},
  pages   = {090701}
}

@article{fukami2019sr,
  author    = {Fukami, Kai and Fukagata, Koji and Taira, Kunihiko},
  title     = {Super-resolution reconstruction of turbulent flows with machine
               learning},
  journal   = {Journal of Fluid Mechanics},
  year      = {2019},
  volume    = {870},
  pages     = {106--120},
  publisher = {Cambridge University Press}
}

@article{fukami2021spatiotemporal,
  author    = {Fukami, Kai and Fukagata, Koji and Taira, Kunihiko},
  title     = {Machine-learning-based spatio-temporal super resolution
               reconstruction of turbulent flows},
  journal   = {Journal of Fluid Mechanics},
  year      = {2021},
  volume    = {909},
  pages     = {A9},
  publisher = {Cambridge University Press}
}

@article{fukami2024singlesnapshot,
  author  = {Fukami, Kai and Fukagata, Koji},
  title   = {Single-snapshot machine learning for super-resolution of
             turbulence},
  journal = {Journal of Fluid Mechanics},
  year    = {2024},
  volume  = {998},
  pages   = {A17}
}

@article{fukami2024single,
  author    = {Fukami, Kai and Taira, Kunihiko},
  title     = {Single-snapshot machine learning for super-resolution of
               turbulence},
  journal   = {Journal of Fluid Mechanics},
  year      = {2024},
  volume    = {1001},
  pages     = {A32},
  publisher = {Cambridge University Press}
}

@article{fukami2023review,
  author  = {Fukami, Kai and Fukagata, Koji and Taira, Kunihiko},
  title   = {Machine-learning-assisted flow reconstruction and
             super-resolution: a perspective},
  journal = {Annual Review of Fluid Mechanics},
  year    = {2023}
}

@article{fukami2023super,
  author    = {Fukami, Kai and Fukagata, Koji and Taira, Kunihiko},
  title     = {Super-resolution analysis via machine learning: a survey for
               fluid flows},
  journal   = {Theoretical and Computational Fluid Dynamics},
  year      = {2023},
  volume    = {37},
  number    = {4},
  pages     = {421--444},
  publisher = {Springer}
}

@article{fukami2025observable,
  author    = {Fukami, Kai and Taira, Kunihiko},
  title     = {Observable-augmented manifold learning for multi-source
               turbulent flow data},
  journal   = {Journal of Fluid Mechanics},
  year      = {2025},
  volume    = {1010},
  pages     = {R4},
  publisher = {Cambridge University Press}
}

@article{liu2020turbsr,
  author  = {Liu, Bo and Tang, Jiupeng and Huang, Hailong and Lu, Xi-Yun},
  title   = {Deep learning methods for super-resolution reconstruction of
             turbulent flows},
  journal = {Physics of Fluids},
  year    = {2020},
  volume  = {32},
  number  = {2},
  pages   = {025105}
}

@article{gao2021picnnsr,
  author  = {Gao, Han and Sun, Luning and Wang, Jian-Xun},
  title   = {Super-resolution and denoising of fluid flow using
             physics-informed convolutional neural networks without
             high-resolution labels},
  journal = {Physics of Fluids},
  year    = {2021},
  volume  = {33},
  number  = {7},
  pages   = {073603}
}

@article{kim2021unsupervised,
  author  = {Kim, H. and others},
  title   = {Unsupervised deep learning for super-resolution reconstruction
             of turbulence},
  journal = {Journal of Fluid Mechanics},
  year    = {2021},
  volume  = {910},
  pages   = {A29}
}

@article{yousif2023reconstruct3d,
  author  = {Yousif, Mahmoud Z. and Yu, Linqi and Lim, Hee-Chang},
  title   = {A deep-learning approach for reconstructing {3D} turbulent flows
             from {2D} observation data},
  journal = {Scientific Reports},
  year    = {2023},
  volume  = {13},
  pages   = {29525}
}

@article{pang2024sgssr,
  author  = {Pang, Z. and others},
  title   = {A deep-learning super-resolution reconstruction model of
             subgrid-scale turbulent flow fields in large-eddy simulation},
  journal = {Computers \& Structures},
  year    = {2024},
  volume  = {298},
  pages   = {107392}
}

@article{page2025dynamicsloss,
  author  = {Page, Jacob and others},
  title   = {Super-resolution of turbulence with dynamics in the loss},
  journal = {Journal of Fluid Mechanics},
  year    = {2025},
  volume  = {1011},
  pages   = {A38}
}

@article{page2025super,
  author    = {Page, Jacob},
  title     = {Super-resolution of turbulence with dynamics in the loss},
  journal   = {Journal of Fluid Mechanics},
  year      = {2025},
  volume    = {1002},
  pages     = {R3},
  publisher = {Cambridge University Press}
}

@article{bao2023deeplearning,
  author  = {Bao, Kai and others},
  title   = {Deep learning method for super-resolution reconstruction of
             turbulent flows: a review or perspective},
  journal = {Journal of Theoretical and Applied Mechanics Letters},
  year    = {2023},
  volume  = {13},
  pages   = {100451}
}

@article{stengel2020adversarial,
  author  = {Stengel, Karen and Glaws, Andrew and Hettinger, Dylan
             and King, Ryan N.},
  title   = {Adversarial super-resolution of climatological wind and solar
             data},
  journal = {Proceedings of the National Academy of Sciences},
  year    = {2020},
  volume  = {117},
  number  = {29},
  pages   = {16805--16815}
}

@article{ozawa2024spatial,
  author    = {Ozawa, Yuta and Honda, Harutaka and Nonomura, Taku},
  title     = {Spatial superresolution based on simultaneous dual {PIV}
               measurement with different magnification},
  journal   = {Experiments in Fluids},
  year      = {2024},
  volume    = {65},
  number    = {4},
  pages     = {42},
  publisher = {Springer}
}

@article{yu2025three,
  author    = {Yu, Linqi and Chen, Yanyun and Yousif, Mustafa Z
               and Lim, Hee-Chang},
  title     = {Three-dimensional super-resolution reconstruction of turbulent
               flow using {3D-ESRGAN} with random sampling strategy},
  journal   = {Computers \& Fluids},
  year      = {2025},
  pages     = {106890},
  publisher = {Elsevier}
}

@article{wu2026super,
  author    = {Wu, Hua-Lin and Xu, Ao and Xi, Heng-Dong},
  title     = {Super-resolution reconstruction of turbulent flows from a
               single {Lagrangian} trajectory},
  journal   = {Journal of Fluid Mechanics},
  year      = {2026},
  volume    = {1026},
  pages     = {A46},
  publisher = {Cambridge University Press}
}

@article{wu2025generalizable,
  author    = {Wu, Haokai and Cao, Yong and Chen, Yaoran and Laima, Shujin
               and Chen, Wen-Li and Zhou, Dai and Li, Hui},
  title     = {Generalizable super-resolution turbulence reconstruction from
               minimal training data},
  journal   = {Journal of Fluid Mechanics},
  year      = {2025},
  volume    = {1024},
  pages     = {A27},
  publisher = {Cambridge University Press}
}

@article{ye2025enhancing,
  author    = {Ye, Mao-kun and Fan, Guo-qing and Wan, De-cheng},
  title     = {Enhancing turbulent channel flow super-resolution via
               physics-guided deep learning with global pressure correlations},
  journal   = {Journal of Hydrodynamics},
  year      = {2025},
  pages     = {1--10},
  publisher = {Springer}
}

@article{stahl2025sparse,
  author    = {Stahl, Spencer L and Benton, Stuart I},
  title     = {Sparse flow reconstruction methods to reduce the costs of
               analyzing large unsteady datasets},
  journal   = {Journal of Computational Physics},
  year      = {2025},
  volume    = {534},
  pages     = {114037},
  publisher = {Elsevier}
}

@article{dong2016srcnn,
  author  = {Dong, Chao and Loy, Chen Change and He, Kaiming and Tang, Xiaoou},
  title   = {Image Super-Resolution Using Deep Convolutional Networks},
  journal = {IEEE Transactions on Pattern Analysis and Machine Intelligence},
  year    = {2016},
  volume  = {38},
  number  = {2},
  pages   = {295--307}
}

@inproceedings{dong2016fsrcnn,
  author    = {Dong, Chao and Loy, Chen Change and Tang, Xiaoou},
  title     = {Accelerating the Super-Resolution Convolutional Neural Network},
  booktitle = {European Conference on Computer Vision},
  year      = {2016},
  pages     = {391--407}
}

@inproceedings{shi2016espcn,
  author    = {Shi, Wenzhe and Caballero, Jose and Husz{\'a}r, Ferenc
               and Totz, Johannes and Aitken, Andrew P. and Bishop, Rob
               and Rueckert, Daniel and Wang, Zehan},
  title     = {Real-Time Single Image and Video Super-Resolution Using an
               Efficient Sub-Pixel Convolutional Neural Network},
  booktitle = {CVPR},
  year      = {2016},
  pages     = {1874--1883}
}

@inproceedings{kim2016vdsr,
  author    = {Kim, Jiwon and Lee, Jung Kwon and Lee, Kyoung Mu},
  title     = {Accurate Image Super-Resolution Using Very Deep Convolutional
               Networks},
  booktitle = {CVPR},
  year      = {2016},
  pages     = {1646--1654}
}

@inproceedings{kim2016drcn,
  author    = {Kim, Jiwon and Lee, Jung Kwon and Lee, Kyoung Mu},
  title     = {Deeply-Recursive Convolutional Network for Image
               Super-Resolution},
  booktitle = {CVPR},
  year      = {2016},
  pages     = {1637--1645}
}

@inproceedings{tai2017drrn,
  author    = {Tai, Ying and Yang, Jian and Liu, Xiaoming},
  title     = {Image Super-Resolution via Deep Recursive Residual Network},
  booktitle = {CVPR},
  year      = {2017},
  pages     = {3147--3155}
}

@inproceedings{lai2017lapsrn,
  author    = {Lai, Wei-Sheng and Huang, Jia-Bin and Ahuja, Narendra
               and Yang, Ming-Hsuan},
  title     = {Deep {Laplacian} Pyramid Networks for Fast and Accurate
               Super-Resolution},
  booktitle = {CVPR},
  year      = {2017},
  pages     = {624--632}
}

@inproceedings{lim2017edsr,
  author    = {Lim, Bee and Son, Sanghyun and Kim, Heewon and Nah, Seungjun
               and Lee, Kyoung Mu},
  title     = {Enhanced Deep Residual Networks for Single Image
               Super-Resolution},
  booktitle = {CVPR Workshops},
  year      = {2017},
  pages     = {1132--1140}
}

@inproceedings{zhang2018rdn,
  author    = {Zhang, Yulun and Tian, Yapeng and Kong, Yu and Zhong, Bineng
               and Fu, Yun},
  title     = {Residual Dense Network for Image Super-Resolution},
  booktitle = {CVPR},
  year      = {2018},
  pages     = {2472--2481}
}

@inproceedings{haris2018dbpn,
  author    = {Haris, Muhammad and Shakhnarovich, Gregory and Ukita, Norimichi},
  title     = {Deep Back-Projection Networks for Super-Resolution},
  booktitle = {CVPR},
  year      = {2018},
  pages     = {1664--1673}
}

@inproceedings{zhang2018rcan,
  author    = {Zhang, Yulun and Li, Kunpeng and Li, Kai and Wang, Lichen
               and Zhong, Bineng and Fu, Yun},
  title     = {Image Super-Resolution Using Very Deep Residual Channel
               Attention Networks},
  booktitle = {ECCV},
  year      = {2018},
  pages     = {286--301}
}

@inproceedings{wang2018esrgan,
  author    = {Wang, Xintao and Yu, Ke and Wu, Shixiang and Gu, Jinjin
               and Liu, Yihao and Dong, Chao and Qiao, Yu
               and Change Loy, Chen},
  title     = {{ESRGAN}: Enhanced Super-Resolution Generative Adversarial
               Networks},
  booktitle = {ECCV Workshops},
  year      = {2018}
}

@inproceedings{ahn2018carn,
  author    = {Ahn, Namhyuk and Kang, Byungkon and Sohn, Kyung-Ah},
  title     = {Fast, Accurate, and Lightweight Super-Resolution with Cascading
               Residual Network},
  booktitle = {ECCV},
  year      = {2018},
  pages     = {252--268}
}

@inproceedings{kong2021classsr,
  author    = {Kong, Xiangtao and Zhao, Haiyang and Qiao, Yu and Dong, Chao},
  title     = {{ClassSR}: A General Framework to Accelerate Super-Resolution
               Networks by Data Characteristic},
  booktitle = {CVPR},
  year      = {2021},
  pages     = {12016--12025}
}

@inproceedings{liang2021swinir,
  author    = {Liang, Jingyun and Cao, Jiezhang and Sun, Guolei and Zhang, Kai
               and Van Gool, Luc and Timofte, Radu},
  title     = {{SwinIR}: Image Restoration Using {Swin} Transformer},
  booktitle = {ICCV Workshops},
  year      = {2021}
}

@inproceedings{saharia2022image,
  author    = {Saharia, Chitwan and Ho, Jonathan and Chan, William
               and Salimans, Tim and Fleet, David J. and Norouzi, Mohammad},
  title     = {Image Super-Resolution via Iterative Refinement},
  booktitle = {IEEE Transactions on Pattern Analysis and Machine Intelligence},
  year      = {2022},
  volume    = {45},
  number    = {4},
  pages     = {4713--4726}
}

@inproceedings{he2016deep,
  author    = {He, Kaiming and Zhang, Xiangyu and Ren, Shaoqing and Sun, Jian},
  title     = {Deep Residual Learning for Image Recognition},
  booktitle = {CVPR},
  year      = {2016},
  pages     = {770--778}
}

@article{vaswani2017attention,
  author  = {Vaswani, Ashish and Shazeer, Noam and Parmar, Niki
             and Uszkoreit, Jakob and Jones, Llion and Gomez, Aidan N.
             and Kaiser, {\L}ukasz and Polosukhin, Illia},
  title   = {Attention Is All You Need},
  journal = {Advances in Neural Information Processing Systems},
  year    = {2017},
  volume  = {30}
}

@article{kingma2014adam,
  author  = {Kingma, Diederik P. and Ba, Jimmy},
  title   = {Adam: A Method for Stochastic Optimization},
  journal = {arXiv preprint arXiv:1412.6980},
  year    = {2014}
}

@incollection{paszke2019pytorch,
  author    = {Paszke, Adam and Gross, Sam and Massa, Francisco
               and Lerer, Adam and Bradbury, James and Chanan, Gregory
               and Killeen, Trevor and Lin, Zeming and Gimelshein, Natalia
               and Antiga, Luca and Desmaison, Alban and K\"{o}pf, Andreas
               and Yang, Edward and DeVito, Zachary and Raison, Martin
               and Tejani, Alykhan and Chilamkurthy, Sasank
               and Steiner, Benoit and Fang, Lu and Bai, Junjie
               and Chintala, Soumith},
  title     = {{PyTorch}: An Imperative Style, High-Performance Deep Learning
               Library},
  booktitle = {Advances in Neural Information Processing Systems},
  year      = {2019},
  volume    = {32}
}

@book{engl1996regularization,
  author    = {Engl, Heinz W. and Hanke, Martin and Neubauer, Andreas},
  title     = {Regularization of Inverse Problems},
  publisher = {Springer},
  year      = {1996},
  address   = {Dordrecht}
}

@book{canuto1988spectral,
  author    = {Canuto, Claudio and Hussaini, M. Yousuff and Quarteroni, Alfio
               and Zang, Thomas A.},
  title     = {Spectral Methods in Fluid Dynamics},
  publisher = {Springer},
  year      = {1988},
  address   = {New York},
  series    = {Springer Series in Computational Physics}
}

@article{mallat1989multiresolution,
  author  = {Mallat, St\'{e}phane G.},
  title   = {A Theory for Multiresolution Signal Decomposition: The Wavelet
             Representation},
  journal = {IEEE Transactions on Pattern Analysis and Machine Intelligence},
  year    = {1989},
  volume  = {11},
  number  = {7},
  pages   = {674--693}
}

@article{price2023probabilistic,
  author  = {Price, Ilan and Sanchez-Gonzalez, Alvaro and Alet, Ferran
             and Ewalds, Timo and others},
  title   = {Probabilistic weather forecasting with machine learning},
  journal = {Nature},
  year    = {2024},
  volume  = {637},
  pages   = {84--90}
}

@book{trottenberg2000multigrid,
  author    = {Trottenberg, Ulrich and Oosterlee, Cornelius W. and Sch\"{u}ller, Anton},
  title     = {Multigrid},
  publisher = {Academic Press},
  year      = {2000},
  address   = {San Diego, CA}
}

@inproceedings{karras2018progressive,
  author    = {Karras, Tero and Aila, Timo and Laine, Samuli and Lehtinen, Jaakko},
  title     = {Progressive Growing of {GAN}s for Improved Quality, Stability, and Variation},
  booktitle = {International Conference on Learning Representations (ICLR)},
  year      = {2018}
}

@inproceedings{bengio2009curriculum,
  author    = {Bengio, Yoshua and Louradour, J\'{e}r\^{o}me and Collobert, Ronan and Weston, Jason},
  title     = {Curriculum Learning},
  booktitle = {Proceedings of the 26th International Conference on Machine Learning (ICML)},
  pages     = {41--48},
  year      = {2009}
}

\end{document}